\PassOptionsToPackage{square,comma,numbers,sort&compress}{natbib}

\documentclass{article}

\usepackage[preprint]{neurips_2022}

\usepackage[utf8]{inputenc} 
\usepackage[T1]{fontenc}    
\usepackage[colorlinks,bookmarks=false]{hyperref}
\usepackage{url}            
\usepackage{booktabs}       
\usepackage{amsfonts}       
\usepackage{nicefrac}       
\usepackage{microtype}      
\usepackage{xcolor}         

\usepackage{multirow}
\usepackage{subfigure}
\usepackage{booktabs}
\usepackage{wrapfig}
\usepackage{graphicx}
\usepackage{amsmath}
\usepackage{bbding}
\usepackage{colortbl} 
\usepackage{makecell}

\newcommand{\tabincell}[2]{\begin{tabular}{@{}#1@{}}#2\end{tabular}}
\newcommand*\samethanks[1][\value{footnote}]{\footnotemark[#1]}

\newcommand{\etal}{ \emph{et al.} }
\newcommand{\eg}{\emph{e.g.}, }
\newcommand{\ie}{ \emph{i.e.}, }

\title{Inception Transformer}

\author{Chenyang Si$^{1}$\thanks{Equal contribution. Weihao Yu did this work during an internship at Sea AI Lab.}~~~Weihao Yu$^{1,2}$\samethanks
~~~Pan Zhou$^{1}$~Yichen Zhou$^{1,2}$~Xinchao Wang$^{2}$~Shuicheng Yan$^{1}$ 
\vspace{3 mm}
\\
\textsuperscript{1}Sea AI Lab ~~~~ \textsuperscript{2}National University of Singapore\\
\tt\small\{sicy,yuweihao,zhoupan,zhouyc,yansc\}@sea.com, xinchao@nus.edu.sg
}

\begin{document}

\maketitle
\begin{abstract}

Recent studies show that Transformer has strong capability of building long-range dependencies, yet is incompetent in capturing high frequencies that predominantly convey local information. To tackle this issue, we present a novel and general-purpose \textit{Inception Transformer}, or \textit{iFormer} for short, that effectively learns comprehensive features with both high- and low-frequency information in visual data. 
Specifically,  we design an Inception mixer to explicitly graft the advantages of convolution and max-pooling for capturing the high-frequency information to Transformers. Different from recent hybrid frameworks, the Inception mixer brings greater efficiency through a channel splitting mechanism to adopt parallel convolution/max-pooling path and self-attention path as high- and low-frequency mixers, while having the flexibility to model discriminative information scattered within a wide frequency range. Considering that bottom layers play more roles in capturing high-frequency details while top layers more in modeling low-frequency global information, we further introduce a frequency ramp structure, \ie gradually decreasing the dimensions fed to the high-frequency mixer and increasing those to the low-frequency mixer, which can effectively trade-off high- and low-frequency components across different layers. We benchmark the iFormer on a series of vision tasks, and showcase that it achieves impressive performance on  image classification, COCO detection and ADE20K segmentation. For example, our iFormer-S hits the top-1 accuracy of $83.4\%$ on ImageNet-1K, much higher than DeiT-S by $3.6\%$, and even slightly better than much bigger model Swin-B ($83.3\%$) with only 1/4 parameters and 1/3 FLOPs. Code and models will be released at \url{https://github.com/sail-sg/iFormer}.
 

\end{abstract}

\section{Introduction}

Transformer~\cite{transformer} has taken the natural language processing (NLP) domain by storm, achieving surprisingly high performance in many NLP tasks, \eg machine translation~\cite{brown2020language} and question-answering~\cite{chowdhery2022palm}.  
This is largely attributed to its strong capability of modeling long-range dependencies in the data with self-attention mechanism.
Its success has led researchers to investigate its adaptation to the computer vision field, and 
Vision Transformer (ViT) \cite{dosovitskiy2020image}  is a pioneer.
This architecture is directly inherited from NLP~\cite{transformer}, but applied to image classification with raw image patches as input. 
Later, many ViT variants~\cite{liu2021swin,pvt,yu2021metaformer,hudson2021generative,arnab2021vivit,beal2020toward,fang2021you,zheng2021rethinking,xie2021segformer}  have been developed to boost performance or scale to a wider range of vision tasks, \eg object detection \cite{beal2020toward,fang2021you} and segmentation \cite{zheng2021rethinking,xie2021segformer}.

\begin{figure}[!t]
	\centering
	\subfigure[ ]{
        \label{fig_idea_and_acc:a}
        \includegraphics[width=0.12\textwidth]{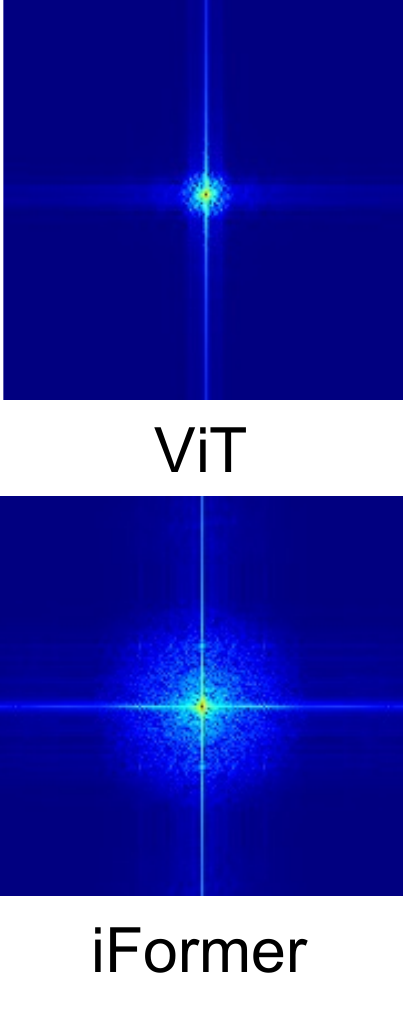}
        }
    \subfigure[ ]{
        \label{fig_idea_and_acc:b}
        \includegraphics[width=0.4\textwidth]{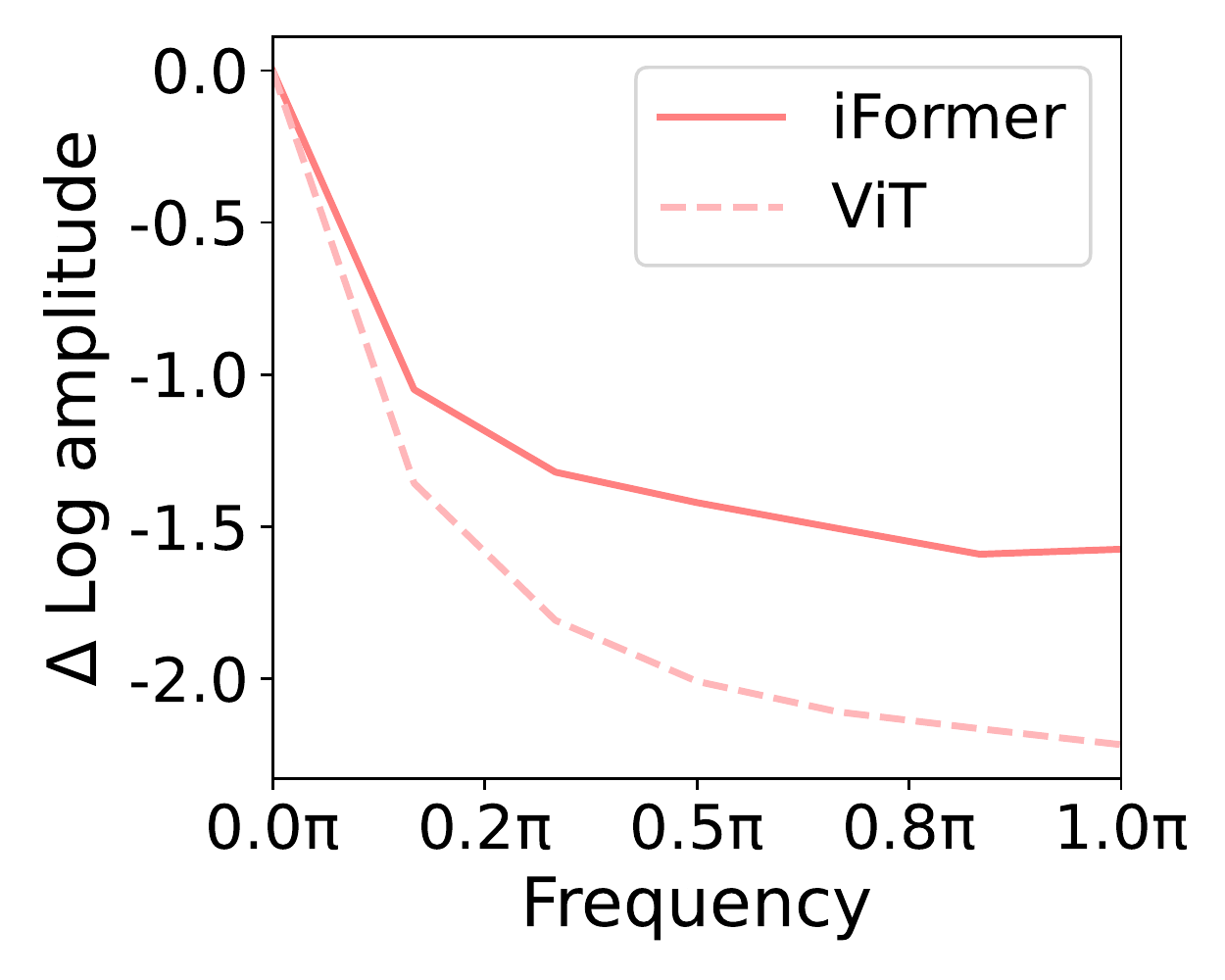}
        }
    \subfigure[ ]{
        \label{fig_idea_and_acc:c}
        \includegraphics[width=0.4\textwidth]{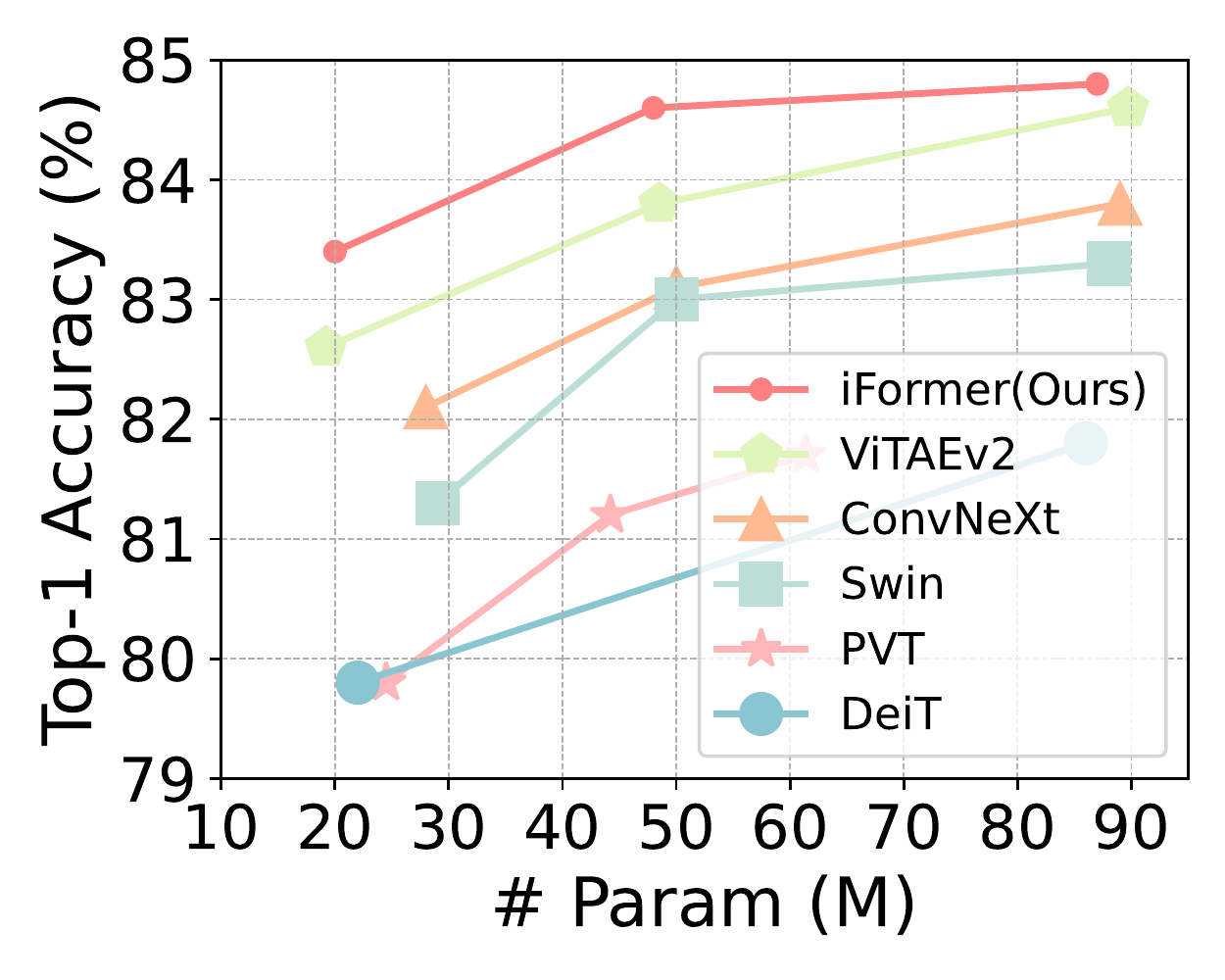}
        }
    \vspace{-2mm}
  	\caption{\textbf{(a) Fourier spectrum of ViT~\cite{vit} and iFormer. (b) Relative log amplitudes of Fourier transformed feature maps. 
  	(c) Performance of models on ImageNet-1K validation set.} (a) and (b) show that iFormer captures more high-frequency signals. 
  	}
  	\label{fig_idea_and_acc}
  	\vspace{-15pt}
\end{figure}

ViT and its variants are highly  capable of capturing low-frequencies in the visual data~\cite{park2021vision}, mainly including global shapes and structures of a scene or object, but are not very powerful for learning high-frequencies, mainly including local edges and textures.  
This can be intuitively explained: 
self-attention, the main operation used in ViTs to exchange information among non-overlap patch tokens, is a  global operation and much more capable of capturing global information  (low frequencies) in the data than local information (high frequencies). 
As shown in Fig.~\ref{fig_idea_and_acc:a} and \ref{fig_idea_and_acc:b}, the Fourier spectrum and relative log amplitudes of the Fourier show that ViT tends to well capture  low-frequency signals but few high-frequency signals. 
This observation also accords with the empirical results in  \cite{park2021vision}, which shows ViT presents the characteristics of low-pass filters.
This low-frequency preferability impairs the performance of ViTs, as 1) low-frequency information filling in all the layers may deteriorate high-frequency components, \eg local textures, and weakens modeling capability of ViTs; 2) high-frequency information is also discriminative and can benefit many tasks, \eg (fine-grained) classification. 
Actually, human visual system extracts visual elementary features at different frequencies \cite{bullier2001integrated,bar2003cortical,kauffmann2014neural}: low frequency provides global information about a visual stimulus, and high frequency conveys local spatial changes in the image (\eg local edges/textures).
Hence, it is necessary to develop a new ViT architecture for capturing both high and low frequencies in the visual data.

CNNs are the most fundamental backbone for general vision tasks. Unlike ViTs, they cover more local information through local convolution within the receptive fields, thus effectively extracting high-frequency representations \cite{wang2020high,yin2019fourier}.
Recent studies \cite{xiao2021early,li2022uniformer,xu2021vitae,dai2021coatnet,cvt} have integrated CNNs and ViTs considering their complementary advantages.
Some methods \cite{xiao2021early,li2022uniformer,dai2021coatnet,cvt} stack convolution and attention layers in a serial manner to inject the local information into global context. 
Unfortunately, this serial manner only models one type of dependency, either global or local, in one layer, and discards the global information during locality modeling, or vice versa. 
Other works \cite{xu2021vitae, xu2021co} adopt parallel attention and convolution to learn global and local dependencies of the input at the same time. 
However, it is found in \cite{raghu2021vision} that part of the channels are for processing local information and the other for global modeling, meaning current parallel structures have information redundancy if processing all channels in each branch.

To address this issue, we propose a simple and efficient \textit{Inception Transformer (iFormer)}, as shown in Fig.~\ref{fig_backbone}, which grafts the merit of CNNs for capturing high-frequencies to ViTs.
The key component in iFormer is an  Inception token mixer as shown in Fig.~\ref{fig_mixer}. 
This Inception mixer aims to augment the perception capability of ViTs in the frequency spectrum by capturing both high and low frequencies in the data. 
To this end, the Inception mixer first  splits the input feature along the channel dimension, and then feeds the split components into high-frequency mixer and low-frequency mixer respectively. 
Here the high-frequency mixer consists of a  max-pooling operation and a parallel convolution operation, while the low-frequency mixer is implemented by a vanilla self-attention in ViTs.
In this way, our iFormer can effectively capture particular frequency information on the corresponding channel, and thus learn more comprehensive features within a wide frequency range compared with vanilla ViTs, which can be clearly observed in Fig.~\ref{fig_idea_and_acc:a} and \ref{fig_idea_and_acc:b}.

Moreover, we find that lower layers often need more local information, while higher layers desire more  global information, which also accords with the observations in~\cite{raghu2021vision}. 
This is because, like in human visual system, the details in high frequency components help lower layers to capture visual elementary features and also to gradually gather local information for having a global understanding of the input.  
Inspired by this, we design a frequency ramp structure.
In particular, from lower to higher layers, we gradually feed more channel dimensions to low-frequency mixer and fewer channel dimensions to high-frequency mixer.
This structure can trade-off  high-frequency and low-frequency components across all layers.
Its effectiveness has been verified by experimental results in Sec.~\ref{exp}.

Experimental results show that  iFormer surpasses state-of-the-art ViTs and CNNs on several vision tasks, including image classification, object detection and segmentation. For example, as shown in Fig.~\ref{fig_idea_and_acc:c}, with different model sizes, iFormer makes consistent improvements over popular frameworks on ImageNet-1K \cite{deng2009imagenet}, \eg DeiT~\cite{deit}, Swin~\cite{liu2021swin} and ConvNeXt \cite{convnext}. 
Meanwhile,  iFormer outperforms recent frameworks on COCO \cite{lin2014microsoft} detection and ADE20K \cite{zhou2017scene} segmentation.  

\section{Related work}
Transformers \cite{transformer} are firstly proposed for machine translation tasks and then become popular in other tasks like natural language understanding \cite{bert, xlnet, roberta} and generation \cite{gpt, gpt3} in NLP domain, as well as image classification \cite{vit, deit, t2t}, object detection \cite{detr, zhu2020deformable, pvt} and semantic segmentation \cite{setr, transunet} in computer vision. 
The attention module in Transformers has an outstanding ability to capture global dependency, but it makes the models produce similar representations across layers \cite{raghu2021vision}.
Moreover, self-attention mainly captures low-frequency information and tends to neglect high-frequency components related to the detailed information \cite{park2021vision}.

CNNs \cite{lecun1998gradient, krizhevsky2012imagenet, simonyan2014very, szegedy2015going, resnet} are the de-facto model for vision tasks due to their outstanding ability to model local dependency \cite{alexnet, vgg, resnet} as well as extract high-frequency \cite{wang2020high}.
With these advantages, CNNs are rapidly introduced into Transformers in a serial or parallel manner \cite{cvt, jiang2021all, dai2021coatnet, tu2022maxvit, chen2021glit, xu2021co, xu2021vitae}. 
For serial methods, convolutions are applied at different positions of the Transformer. CvT \cite{cvt} and PVT-v2 \cite{pvtv2} replace the hard patch embedding with a layer of overlapping convolution. LV-ViT \cite{jiang2021all}, LeViT \cite{graham2021levit} and ViT$_C$ \cite{xiao2021early} further stack several layers of convolutions as the stem for models, which is found helpful in training and achieving better performance. Besides the stem, ViT-hybrid \cite{vit}, CoAtNet~\cite{dai2021coatnet}, Hybrid-MS \cite{zhao2021battle} and UniFormer \cite{li2022uniformer} design early stages with convolution layers.
However, the combination of convolution and attention in a serial order means each layer can only process either high or low frequency and neglects the other part.  To enable each layer to process different frequencies, we adopt the parallel manner to combine convolution and attention in a token mixer.

Compared with serial methods, there are not many works combining attention and convolution in a parallel manner in literature. 
CoaT \cite{xu2021co} and ViTAE \cite{xu2021vitae} introduce convolution as a branch parallel to attention and utilize elementwise sum to merge the output of the two branches.
However, Raghu \etal find that some channels tend to extract local dependency while others are for modeling global information \cite{raghu2021vision}, indicating redundancy for the current parallel mechanism to process all channels in different branches. In contrast, we split channels into branches of high and low frequencies. GLiT \cite{chen2021glit} also adopt parallel manner but it directly concatenate the features from convolution and attention branches as the mixer output, lacking the fusion of features in different frequencies. Instead, we design a explicit fusion module to merge the outputs from low- and high-frequency branches.

\section{Method}
\subsection{Revisit Vision Transformer}

We first revisit the Vision Transformer. 
For vision tasks, Transformers first split the input image into a sequence of tokens, and each patch token is projected into a hidden representation vector with a leaner layer, denoted as $\{\boldsymbol{x}_1,\boldsymbol{x}_2,...,\boldsymbol{x}_N\}$ or $\boldsymbol{X} \in \mathbb{R}^{N \times C}$, where $N$ is the number of patch tokens and $C$ indicates the dimension of features. Then, all of the tokens are combined with a positional embedding and fed into the Transformer layers that contain multi-head self-attention (MSA) and a feed-forward network (FFN).

\begin{figure}[t]
	\centering
	\includegraphics[width=0.98\textwidth]{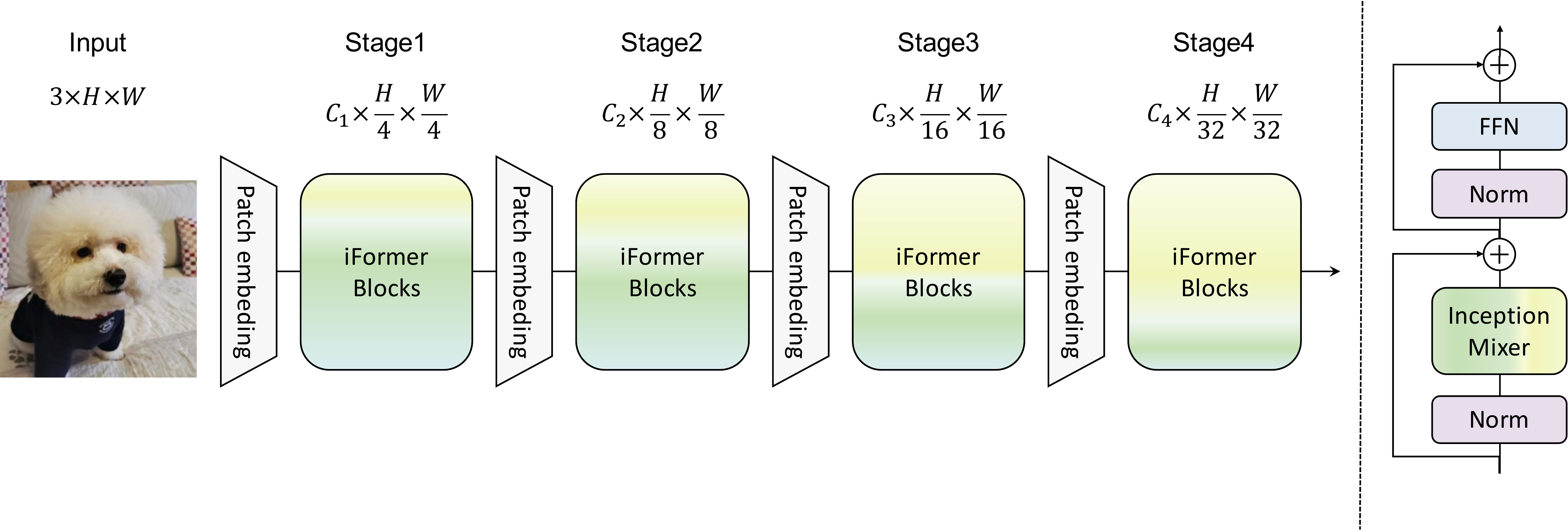}
	\vspace{-2mm}
	\caption{\textbf{The overall architecture of iFormer and details of iFormer block }. For each block, yellow and green indicate low- and high-frequency information, respectively. Best viewed in color.}
	\label{fig_backbone}
\end{figure}

In MSA, the attention-based mixer exchanges information between all patch tokens so that it strongly focuses on aggregating the global dependency across all layers. 
However, excessive propagation of global information would strengthen the low-frequency representation. 
It can be seen from the visualization of Fourier spectrum in Fig.~\ref{fig_idea_and_acc:a} that low-frequency information dominates the representations of ViT \cite{vit}.
This actually impairs the performance of ViTs, as it may deteriorate the high-frequency components, \eg local textures, and weakens the modeling capability of ViTs \cite{park2021vision}.
In the visual data, high-frequency information is also discriminative and can benefit many tasks \cite{wang2020high,yin2019fourier}. Hence, to address the issue, we propose a simple and efficient Inception Transformer, as shown in Fig.~\ref{fig_backbone}, with two key novelties, \ie~Inception mixer and frequency ramp structure.

\subsection{Inception token mixer}
\begin{wrapfigure}{R}{0.3\textwidth}
  \centering
    \includegraphics[width=0.85\linewidth, height=1.2\linewidth]{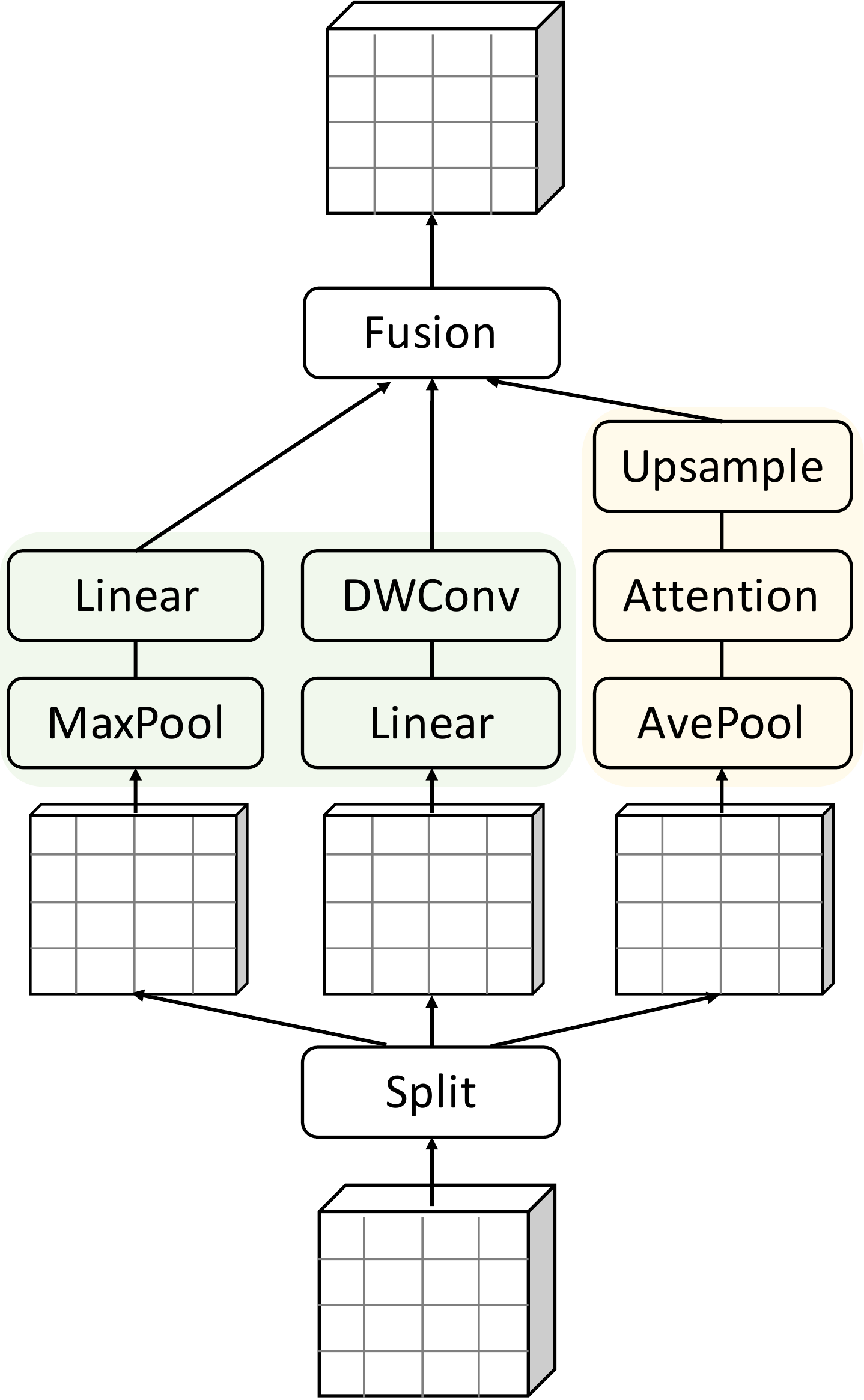}
 \caption{\textbf{The details of Inception mixer.}}
  \label{fig_mixer}
  \vspace{-20pt}
\end{wrapfigure}
We propose an Inception mixer to graft the powerful capability of CNNs for extracting high-frequency representation to Transformers. 
Its detailed architecture is depicted in Fig.~\ref{fig_mixer}. We use the name of ``Inception" since the token mixer is highly inspired by the Inception module \cite{szegedy2015going, ioffe2015batch, szegedy2016rethinking, szegedy2017inception} with multiple branches.
Instead of directly feeding image tokens into the MSA mixer, the Inception mixer first splits the input feature along the channel dimension, and then respectively feeds the split components into high-frequency mixer and low-frequency mixer. Here the high-frequency mixer consists of a max-pooling operation and a parallel convolution operation, while the low-frequency mixer is implemented by a self-attention.

Technically, given the input feature map $\boldsymbol{X} \in \mathbb{R}^{N \times C}$, it is factorized $\boldsymbol{X}$ into $\boldsymbol{X}_h \in \mathbb{R}^{N \times C_h}$ and $\boldsymbol{X}_l \in \mathbb{R}^{N \times C_l} $  along the channel dimension, where $C_h + C_l = C$. Then, $\boldsymbol{X}_h$ and $\boldsymbol{X}_l$ are assigned to high-frequency mixer  and low-frequency mixer respectively.

\vspace{-5pt}
\paragraph{High-frequency mixer.} Considering the sharp sensitiveness of the maximum filter and the detail perception of convolution operation, we propose a parallel structure to learn the high-frequency components. We divide the input $\boldsymbol{X}_h$ into $\boldsymbol{X}_{h1} \in \mathbb{R}^{N \times \frac{C_h}{2}}$ and $\boldsymbol{X}_{h2} \in \mathbb{R}^{N \times \frac{C_h}{2}}$ along the channel. As shown in Fig.~\ref{fig_mixer}, $\boldsymbol{X}_{h1}$ is embedded with a max-pooling and a linear layer \cite{szegedy2015going}, and $\boldsymbol{X}_{h2}$ is fed into a linear and a depthwise convolution layer \cite{chollet2017xception, mamalet2012simplifying, sandler2018mobilenetv2}:
\begin{align}
    \label{eqn_high}
    \boldsymbol{Y}_{h1} &=   \mathrm{FC} \left(  \mathrm{MaxPool} \left(  \boldsymbol{X}_{h1}  \right) \right)  \\
    \boldsymbol{Y}_{h2} &=  \mathrm{DwConv} \left(  \mathrm{FC} \left(  \boldsymbol{X}_{h2}  \right) \right) ,
\end{align}
where $\boldsymbol{Y}_{h1}$ and $\boldsymbol{Y}_{h2}$ denote the outputs of high-frequency mixers. 

Finally, the outputs of low- and high-frequency mixers are concatenated along the channel dimension:
\begin{align}
    \label{eqn_concat}
    \boldsymbol{Y_c} &= \mathrm{Concat} \left( \boldsymbol{Y}_{l}, \boldsymbol{Y}_{h1}, \boldsymbol{Y}_{h2}  \right).
\end{align}
The upsample operation in Eq.~(\ref{eqn_attn}) selects the value of the nearest point for each position to be interpolated regardless of any other points, which results in excessive smoothness between adjacent tokens.
We design a fusion module to elegantly overcome this issue, \ie a depthwise convolution exchanging information between patches, while keeping a cross-channel linear layer that works per location like in previous Transformers. 
The final output can be expressed as
\begin{align}
    \label{eqn_fusion}
    \boldsymbol{Y} &= \mathrm{FC} \left( \boldsymbol{Y_c} + \mathrm{DwConv} \left(  \boldsymbol{Y_c}  \right) \right).
\end{align}
Like the vanilla Transformer, our iFormer is equipped with a feed-forward network (FFN), and differently it also incorporates the above Inception token mixer (ITM); LayerNorm (LN) is applied before ITM and FFN. Hence the Inception Transformer block is formally defined as
\begin{align}
    \label{eqn_Inception_block}
    \boldsymbol{Y} &=  \boldsymbol{X} + \mathrm{ITM} \left( \mathrm{LN} \left( \boldsymbol{X} \right) \right) \\
    \boldsymbol{H} &=  \boldsymbol{Y} + \mathrm{FFN} \left( \mathrm{LN} \left( \boldsymbol{Y} \right) \right).
\end{align}

\paragraph{Low-frequency mixer.} 
We use the vanilla multi-head self-attention to communicate information among all tokens for the low-frequency mixer. 
Despite the strong capability of the attention for learning global representation, the large resolution of feature maps would bring large computation cost in lower layers.
We therefore simply utilize an average pooling layer to reduce the spatial scale of $\boldsymbol{X}_l$ before the attention operation and an upsample layer to recover the original spatial dimension after the attention.
This design largely reduces the computational overhead and makes the attention operation focus on embedding global information. This branch can be defined as
\begin{align}
    \label{eqn_attn}
    \boldsymbol{Y}_l &= \mathrm{Upsample} \left(  \mathrm{MSA} \left( \mathrm{AvePooling} \left( \boldsymbol{X}_l  \right) \right) \right),
\end{align}
where $\boldsymbol{Y}_l$ is the output of low-frequency mixer. Note that the kernel size and stride for the pooling and upsample layers are set to 2 only at the first two stages.

\subsection{Frequency ramp structure}

In the general visual frameworks, bottom layers play more roles in capturing high-frequency details while top layers more in modeling low-frequency global information, \ie the hierarchical representations of ResNet~\cite{resnet}. 
Like humans, by capturing the details in high frequency components, lower layers can capture visual elementary features, and also gradually gather local information to achieve a global understanding of the input. 
We are inspired to design a frequency ramp structure which gradually splits  more channel dimensions from lower to higher layers to low-frequency mixer and thus leave fewer channel dimensions to high-frequency mixer. 
Specifically, as shown in Fig.~\ref{fig_backbone}, our backbone has four stages with different channel and spatial dimensions.
For each blocks, we define a channel ratio to better balance the high-frequency and low frequency components, \ie $\frac{C_h}{C}$ and $\frac{C_l}{C}$, where $\frac{C_h}{C} + \frac{C_l}{C} =1$. In the proposed frequency ramp structure, $\frac{C_h}{C}$ gradually decreases from shallow to deep layers, while $\frac{C_l}{C}$ gradually increases. Hence, with the flexible frequency ramp structure, iFormer can effectively trade-off high- and low-frequency components across all layers. The configuration of different iFormer models will be described in the appendix.

\begin{table}[!t]
    \begin{center}
    \caption{ Comparison of different types of models on ImageNet-1K~\cite{deng2009imagenet}.}
    \label{imagenet_sota}
    \footnotesize{
        \begin{tabular}{c|c|l|cccc|cc}
        \toprule 
         \multirow{2}{*}{Model Size} & \multirow{2}{*}{Arch.} & \multirow{2}{*}{Method} & \#Param. & FLOPs & \multicolumn{2}{c}{Input Size} & \multicolumn{2}{|c}{ImageNet} \\
        ~ & ~ & ~ & (M)  & (G) & Train & Test & Top-1 & Top-5  \\
        \midrule
         \multirow{15}{*}{\rotatebox{90}{\tabincell{c}{small model size \\ ($\sim$20M)}}}  &  \multirow{2}{*}{CNN}  
        & RSB-ResNet-50 \cite{resnet, resnetrsb} & 26 & 4.1 & 224 & 224 & 80.4 & - \\
        & & ConvNeXt-T \cite{convnext} & 28 & 4.5 & 224 & 224 & 82.1 & -\\
        \cmidrule{2-9}
        &\multirow{6}{*}{ViT} & Deit-S \cite{deit}                  & 22 & 4.6 & 224 & 224 & 79.8 & 95.0 \\
        ~   &                 & PVT-S \cite{pvt}                                    & 25 & 3.8 & 224 & 224 & 79.8 & - \\
        ~      &              & T2T-14 \cite{t2t}                                   & 22 & 5.2 & 224 & 224 & 80.7 & - \\
        ~         &           & Swin-T \cite{liu2021swin}                           & 29 & 4.5 & 224 & 224 & 81.3 & 95.5 \\
        ~            &        & Focal-T \cite{yang2021focal}                        & 29 & 4.9 & 224 & 224 & 82.2 & 95.9 \\
        ~               &     & CSwin-T \cite{dong2021cswin}                        & 23 & 4.3 & 224 & 224 & 82.7 & - \\
        \cmidrule{2-9}
        &\multirow{7}{*}{Hybrid} & CvT-13 \cite{cvt}               & 20 & 4.5 & 224 & 224 & 81.6 & - \\
        ~  &                  & CoAtNet-0 \cite{dai2021coatnet}                    & 25 & 4.2 & 224 & 224 & 81.6 & - \\
        ~    &                & Container \cite{lu2021container}                   & 22 & 8.1 & 224 & 224 & 82.7 & - \\
        ~      &              & ViTAE-S \cite{xu2021vitae}                         & 24 & 5.6 & 224 & 224 & 82.0 & 95.9 \\
        ~        &            & ViTAEv2-S \cite{zhang2022vitaev2}                  & 19 & 5.7 & 224 & 224 & 82.6 & 96.2 \\
        ~          &          & UniFormer-S \cite{li2022uniformer}                 & 22 & 3.6 & 224 & 224 & 82.9 & - \\
        \rowcolor{gray!15} ~        &            & \textbf{iFormer-S} & \textbf{20} & \textbf{4.8} & \textbf{224} & \textbf{224} & \textbf{83.4} & \textbf{96.6} \\
        \toprule 
        \multirow{13}{*}{\rotatebox{90}{\tabincell{c}{medium model size \\ ($\sim$50M)}}} &  \multirow{3}{*}{CNN}  
        & RSB-ResNet-101 \cite{resnet, resnetrsb} & 45 & 7.9 & 224 & 224 & 81.5 & - \\
        ~& & RSB-ResNet-152 \cite{resnet, resnetrsb} & 60 & 11.6 & 224 & 224 & 82.0 & - \\
        ~& & ConvNeXt-S \cite{convnext} & 50 & 8.7 & 224 & 224 & 83.1 & - \\
        \cmidrule{2-9}
        &\multirow{5}{*}{ViT} & PVT-L \cite{pvt}                      & 61 & 9.8 & 224 & 224 & 81.7 & - \\
        ~&                    & T2T-24 \cite{t2t}                                     & 64 & 13.2 & 224 & 224 & 82.2 & - \\
        ~&                    & Swin-S \cite{liu2021swin}                             & 50 & 8.7 & 224 & 224 & 83.0 & 96.2 \\
        ~&                    & Focal-S \cite{yang2021focal}                          & 51 & 9.1 & 224 & 224 & 83.5 & 96.2	 \\
        ~&                    & CSwin-S \cite{dong2021cswin}                          & 35 & 6.9 & 224 & 224 & 83.6 & - \\
        \cmidrule{2-9}
        &\multirow{5}{*}{Hybrid}  & CvT-21 \cite{cvt}                 & 32 & 7.1 & 224 & 224 & 82.5 & - \\
        ~&                    & CoAtNet-1 \cite{dai2021coatnet}                       & 42 & 8.4 & 224 & 224 & 83.3 & - \\
        ~&                    & ViTAEv2-48M \cite{zhang2022vitaev2}                   & 49 & 13.3 & 224 & 224 & 83.8 & 96.6 \\
        ~&                    & UniFormer-B \cite{li2022uniformer}                    & 50 & 8.3 & 224 & 224 & 83.9 & - \\
        \rowcolor{gray!15} ~&                    & \textbf{iFormer-B} & \textbf{48} & \textbf{9.4} & \textbf{224} & \textbf{224} & \textbf{84.6} & \textbf{97.0} \\
        \toprule 
        \multirow{10}{*}{\rotatebox{90}{\tabincell{c}{large model size \\ ($\sim$100M)}}}  &\multirow{2}{*}{CNN}  
        & RegNetY-16GF \cite{regnet, deit} & 84 & 16.0 & 224 & 224 & 82.9 & -  \\
        & & ConvNeXt-B \cite{convnext} & 89 & 15.4 & 224 & 224 & 83.8 & - \\
        \cmidrule{2-9}
        &\multirow{4}{*}{ViT} & DeiT-B \cite{deit}                  & 86 & 17.5 & 224 & 224 & 81.8 & 95.6 \\
        ~&                    & Swin-B \cite{liu2021swin}                           & 88 & 15.4 & 224 & 224 & 83.3 & 96.5 \\
        ~&                    & Focal-B \cite{yang2021focal}                        & 90 & 16.0 & 224 & 224 & 83.8 & 96.5 \\
        ~&                    & CSwin-B \cite{dong2021cswin}                        & 78 & 15.0 & 224 & 224 & 84.2 & - \\ 
        \cmidrule{2-9}
        &\multirow{4}{*}{Hybrid}  & BoTNet-T7 \cite{srinivas2021bottleneck}        & 79 & 19.3 & 256 & 256 & 84.2 & - \\
        ~&                    & CoAtNet-3 \cite{dai2021coatnet}                     & 168 & 34.7 & 224 & 224 & 84.5 & - \\
        ~&                    & ViTAEv2-B \cite{zhang2022vitaev2}                   & 90 & 24.3 & 224 & 224 & 84.6 & 96.9 \\
        \rowcolor{gray!15} ~&                    & \textbf{iFormer-L} & \textbf{87} & \textbf{14.0} & \textbf{224} & \textbf{224} & \textbf{84.8} & \textbf{97.0} \\
        \bottomrule
    \end{tabular}
    }
    \end{center}
\end{table}

\section{Experiments}\label{exp}
We evaluate our iFormer on several vision benchmark tasks, 
\ie image classification, object detection and semantic segmentation, by comparing it with representative ViTs, CNNs and their hybrid variants. Ablation analysis is also conducted to show the contribution of each novelty in our method. More results will be reported in the appendix.

\subsection{Results on image classification}
\label{exp_1k}
\noindent{\textbf{Setup.}}   
For image classification, we evaluate iFormer on the ImageNet dataset~\cite{deng2009imagenet}.
We train the iFormer model with the standard procedure in~\cite{deit, pvt, li2022uniformer}.
Specifically, we use AdamW optimizer with an initial learning rate $1 \times 10^{-3}$ via cosine decay~\cite{loshchilov2016sgdr}, a momentum of 0.9, and a weight decay of 0.05.  We set the training epoch number as  300 and the input size as 224 $\times$ 224.  We adopt the same data augmentations and regularization methods in DeiT~\cite{deit} for fair comparison. 

We also use LayerScale~\cite{touvron2021going} to train deep models. Like previous studies~\cite{liu2021swin,zhang2022vitaev2}, we further fine tune iFormer on the input size of $384 \times 384$, with the weight decay of $1 \times 10^{-8}$, learning rate of $1 \times 10^{-5}$, batch size of 512. For fairness, we adopt Timm~\cite{timm} to implement and train iFormer. 

\begin{table}[!t]
    \begin{center}
    \caption{Fine-tuning Results with larger resolution (384 $\times$ 384) on ImageNet-1K~\cite{deng2009imagenet}. The models in \textcolor{gray!70}{gray} color are trained with larger input size.}
    \label{imagenet_sota_384}
    \small
    \begin{tabular}{l|cccc|c}
        \toprule 
        \multirow{2}{*}{Method} & \#Param. & FLOPs & \multicolumn{2}{c}{Input Size} & ImageNet \\
         ~ & (M)  & (G) & Train & Test & Top-1   \\
        \midrule
        \textcolor{gray!70}{EfficientNet-B5} \cite{tan2019efficientnet} & \textcolor{gray!70}{30} & \textcolor{gray!70}{9.9} & \textcolor{gray!70}{456} & \textcolor{gray!70}{456} & \textcolor{gray!70}{83.6} \\
        \textcolor{gray!70}{EfficientNetV2-S} \cite{tan2021efficientnetv2} & \textcolor{gray!70}{22} & \textcolor{gray!70}{8.5} & \textcolor{gray!70}{384} & \textcolor{gray!70}{384} & \textcolor{gray!70}{83.9} \\
        CSwin-T$\uparrow$384 \cite{dong2021cswin}        & 23 & 14.0 & 224 & 384 & 84.3  \\
        CvT-13$\uparrow$384 \cite{cvt}                   & 20 & 16.3 & 224 & 384 & 83.0  \\
        CoAtNet-0$\uparrow$384 \cite{dai2021coatnet}     & 20 & 13.4 & 224 & 384 & 83.9 \\
        ViTAEv2-S$\uparrow$384 \cite{zhang2022vitaev2}   & 19 & 17.8 & 224 & 384 & 83.8 \\
        \rowcolor{gray!15} \textbf{iFormer-S$\uparrow$384 } & \textbf{20} & \textbf{16.1} & \textbf{224} & \textbf{384} & \textbf{84.6} \\
        \midrule
        \textcolor{gray!70}{EfficientNet-B7} \cite{tan2019efficientnet} & \textcolor{gray!70}{66} & \textcolor{gray!70}{39.2} & \textcolor{gray!70}{600} & \textcolor{gray!70}{600} & \textcolor{gray!70}{84.3} \\
        \textcolor{gray!70}{EfficientNetV2-M} \cite{tan2021efficientnetv2} & \textcolor{gray!70}{54} & \textcolor{gray!70}{25.0} & \textcolor{gray!70}{480} & \textcolor{gray!70}{480} & \textcolor{gray!70}{85.1} \\
        ViTAEv2-48M $\uparrow$384 \cite{zhang2022vitaev2}         & 49 & 41.1 & 224 & 384 & 84.7 \\
        CSwin-S$\uparrow$384 \cite{dong2021cswin}                 & 35 & 22.0 & 224 & 384 & 85.0  \\
        CoAtNet-1$\uparrow$384  \cite{dai2021coatnet}             & 42 & 27.4 & 224 & 384 & 85.1  \\
        \rowcolor{gray!15} \textbf{iFormer-B$\uparrow$384} & \textbf{48} & \textbf{30.5} & \textbf{224} & \textbf{384} & \textbf{85.7} \\
        \midrule
        \textcolor{gray!70}{EfficientNetV2-L} \cite{tan2021efficientnetv2} & \textcolor{gray!70}{121} & \textcolor{gray!70}{53} & \textcolor{gray!70}{480} & \textcolor{gray!70}{480} & \textcolor{gray!70}{85.7} \\
        Swin-B$\uparrow$384 \cite{liu2021swin}                  & 88 & 47.0 & 224 & 384 & 84.2  \\
        CSwin-B$\uparrow$384 \cite{dong2021cswin}               & 78 & 47.0 & 224 & 384 & 85.4  \\ 
        ViTAEv2-B$\uparrow$384 \cite{zhang2022vitaev2}          & 90 & 74.4 & 224 & 384 & 85.3  \\
        CoAtNet-2$\uparrow$384 \cite{dai2021coatnet}           & 75 & 49.8 & 224 & 384 & 85.7  \\
        \rowcolor{gray!15} \textbf{iFormer-L$\uparrow$384} & \textbf{87} & \textbf{45.3} & \textbf{224} & \textbf{384} & \textbf{85.8}  \\
        \bottomrule
    \end{tabular}
    \end{center}
\end{table}

\noindent{\textbf{Results.}} Table~\ref{imagenet_sota} summarizes the image classification accuracy of all compared methods on  ImageNet. For the small model size ($\sim$20M), our iFormer surpasses both the SoTA ViTs and hybrid ViTs, although some ViTs, \eg Swin~\cite{liu2021swin}, Focal~\cite{yang2021focal} and CSwin~\cite{dong2021cswin}, actually already introduce convolution-like inductive bias into their architectures, and hybrid ViTs directly integrate convolution into ViTs.
Specifically, our iFormer-S respectively gains $0.7\%$ and $0.5\%$ top-1 accuracy advantage over SoTA ViTs (\ie CSwin-T) and hybrid ViTs (\ie UniFormer-S), while enjoying the same or smaller model size.  

For the medium model size ($\sim$50M), iFormer-B achieves 84.6\%  top-1 accuracy, and improves over the SoTA ViTs  and hybrid ViTs with similar model sizes by significant margins 1.0\% and 0.7\% respectively. 
For CNNs, similar to comparison results on medium model size,  our iFormer-B outperforms ConvNeXt-S by $1.5\%$. As for the large mode ($\sim$100M), one can observe similar results on small and medium model sizes. 

Table~\ref{imagenet_sota_384} reports the fine-tuning accuracy on the larger resolution, \ie $384 \times 384$. 
One can observe that iFormer consistently outperforms the counterparts by a significant margin across  different computation settings. These results clearly demonstrate the advantages of iFormer on image classifications.

\begin{table}[!t]
    \begin{center}
    \caption{Performance of object detection and instance segmentation on COCO val2017 \cite{lin2014microsoft}. $AP^b$ and $AP^m$ represent bounding box AP and mask AP, respectively. All models are based on Mask R-CNN~\cite{he2017mask} and trained by 1$\times$ training schedule. The FLOPs are measured at resolution 800$\times$1280. }
    \label{coco_sota_mrcnn}
    \small
    \begin{tabular}{l|c|c|ccc|ccc}
        \toprule 
         \multirow{2}{*}{Method} & \#Param. & FLOPs & \multicolumn{6}{c}{Mask R-CNN 1 $\times$}  \\
        ~ & (M)  & (G) & $AP^b$ & $AP^b_{50}$ & $AP^b_{70}$ & $AP^m$ & $AP^m_{50}$ & $AP^m_{75}$  \\
        \midrule
        ResNet50 \cite{resnet}            & 44 & 260 & 38.0 & 58.6 & 41.4 & 34.4 & 55.1 & 36.7  \\
        PVT-S \cite{pvt}                  & 44 & 245 & 40.4 & 62.9 & 43.8 & 37.8 & 60.1 & 40.3  \\
        TwinsP-S \cite{chu2021twins}      & 44 & 245 & 42.9 & 65.8 & 47.1 & 40.0 & 62.7 & 42.9  \\
        Twins-S \cite{chu2021twins}       & 44 & 228 & 43.4 & 66.0 & 47.3 & 40.3 & 63.2 & 43.4  \\
        Swin-T \cite{liu2021swin}         & 48 & 264 & 42.2 & 64.6 & 46.2 & 39.1 & 61.6 & 42.0  \\
        ViL-S \cite{zhang2021multi}       & 45 & 218 & 44.9 & 67.1 & 49.3 & 41.0 & 64.2 & 44.1  \\
        Focal-T \cite{yang2021focal}      & 49 & 291 & 44.8 & 67.7 & 49.2 & 41.0 & 64.7 & 44.2  \\
        UniFormer-S$_{h14}$ \cite{li2022uniformer} & 41 & 269 & 45.6 & 68.1 & 49.7 & 41.6 & 64.8 & 45.0  \\
        \rowcolor{gray!15} \textbf{iFormer-S}          & \textbf{40} & \textbf{263} & \textbf{46.2} & \textbf{68.5} & \textbf{50.6} & \textbf{41.9} & \textbf{65.3} & \textbf{45.0}  \\
        \midrule
        ResNet101 \cite{resnet}           & 63 & 336 & 40.4 & 61.1 & 44.2 & 36.4 & 57.7 & 38.8  \\
        X101-32                          & 63 & 340 & 41.9 & 62.5 & 45.9 & 37.5 & 59.4 & 40.2  \\
        PVT-M \cite{pvt}                 & 64 & 302 & 42.0 & 64.4 & 45.6 & 39.0 & 61.6 & 42.1  \\
        TwinsP-B \cite{chu2021twins}     & 64 & 302 & 44.6 & 66.7 & 48.9 & 40.9 & 63.8 & 44.2  \\
        Twins-B \cite{chu2021twins}      & 76 & 340 & 45.2 & 67.6 & 49.3 & 41.5 & 64.5 & 44.8  \\
        Swin-S \cite{liu2021swin}        & 69 & 354 & 44.8 & 66.6 & 48.9 & 40.9 & 63.4 & 44.2  \\
        Focal-S \cite{yang2021focal}     & 71 & 401 & 47.4 & 69.8 & 51.9 & 42.8 & 66.6 & 46.1  \\
        CSWin-S  \cite{dong2021cswin}    & 54 & 342 & 47.9 & 70.1 & 52.6 & 43.2 & 67.1 & 46.2  \\
        UniFormer-B \cite{li2022uniformer}  & 69 & 399 & 47.4 & 69.7 & 52.1 & 43.1 & 66.0 & 46.5  \\
        \rowcolor{gray!15} \textbf{iFormer-B}          & \textbf{67} & \textbf{351} & \textbf{48.3} & \textbf{70.3} & \textbf{53.2} & \textbf{43.4} & \textbf{67.2} & \textbf{46.7}  \\
        \bottomrule
    \end{tabular}
    \end{center}
\end{table}

\subsection{Results on  object detection and instance segmentation}

\noindent{\textbf{Setup.}}  
We evaluate iFormer on the COCO object detection and instance segmentation tasks \cite{lin2014microsoft}, where the models are trained on 118K images and evaluated on validation set with 5K images. Here, we use iFormer as the backbone in Mask R-CNN~\cite{he2017mask}. In the training phase, we use iFormer pretrained on ImageNet to initialize the  detector,  
and adopt AdamW to train with an initial learning rate of $1\times 10^{-4}$, a batch size of 16, and 1$\times$ training schedule with 12 epochs.  For training, the input images are resized to be 800 pixels on the shorter side an no more than 1,333 pixels on the longer side. For the test image, its shorter side is fixed to 800 pixels. All experiments are implemented on mmdetection~\cite{mmdetection} codebase.

\noindent{\textbf{Results.}} Table~\ref{coco_sota_mrcnn} reports the box mAP (AP$^b$) and mask mAP (AP$^m$) of the compared models. Under similar computation configurations, iFormers  outperforms all previous backbones. Specifically, compared with popular ResNet~\cite{resnet} backbones, our iFormer-S brings $8.2$ points of AP$^b$ and $7.5$ points AP$^m$ improvements over ResNet50. Compared with various Transformer backbones, our iFormers still maintain the performance superiority over their results. For example, our iFormer-B surpasses UniFormer-B~\cite{li2022uniformer}, Swin-S~\cite{liu2021swin} by $0.9$ points of AP$^b$ and $3.5$ points of AP$^b$ respectively. 
\begin{wraptable}{r}{0.5\textwidth}
    \begin{center}
    \vspace{20pt}
    \caption{Semantic segmentation with semantic FPN~\cite{kirillov2019panoptic} on ADE20K \cite{zhou2017scene}. The FLOPs are measured at resolution 512$\times$2048.}
     \label{coco_sota_seg}
    \small
    \begin{tabular}{l|c|c|c}
        \toprule 
         \multirow{2}{*}{Method} & \#Param. & FLOPs &  mIoU \\
        ~ & (M)  & (G) & (\%)  \\
        \midrule
        ResNet50 \cite{resnet}             & 29 & 183 & 36.7 \\
        PVT-S \cite{pvt}                   & 28 & 161 & 39.8 \\
        TwinsP-S \cite{chu2021twins}       & 28 & 162 & 44.3 \\
        Twins-S \cite{chu2021twins}        & 28 & 144 & 43.2 \\
        Swin-T \cite{liu2021swin}          & 32 & 182 & 41.5 \\
        UniFormer-S$_{h32}$ \cite{li2022uniformer} &25 & 199 & 46.2 \\
        UniFormer-S \cite{li2022uniformer} &25  & 247 & 46.6 \\
        \textcolor{gray!70}{UniFormer-B} \cite{li2022uniformer}  & \textcolor{gray!70}{54} & \textcolor{gray!70}{471} & \textcolor{gray!70}{48.0} \\
        \rowcolor{gray!15} \textbf{iFormer-S}          & \textbf{24} & \textbf{181} & \textbf{48.6} \\
        \bottomrule
    \end{tabular}
    \end{center}
    \vspace{-30pt}
\end{wraptable}

\subsection{Results on semantic segmentation}

\noindent{\textbf{Setup.}} We further evaluate the generality of iFormer through a challenging scene parsing benchmark on semantic segmentation, \ie ADE20K \cite{zhou2017scene}. 
The dataset contains 20K training images and 2K validation images.
We adopt iFormer pretrained on ImageNet as the backbone of the Semantic FPN~\cite{kirillov2019panoptic} framework.  
Following PVT~\cite{pvt} and UniFormer~\cite{li2022uniformer}, we use AdamW with an initial learning rate of $2\times 10^{-4}$ with cosine learning rate schedule to train 80k iterations.   
All experiments are implemented on mmsegmentation~\cite{mmseg2020} codebase.

\noindent{\textbf{Results.}}  
In Table~\ref{coco_sota_seg}, we report the mIoU results of different backbones. On the Semantic FPN~\cite{kirillov2019panoptic} framework, our iFormer consistently outperforms previous backbones on this task, including CNNs and (hybrid) ViTs. For instance, iFormer-S achieves $48.6$ mIoU, surpassing UniFormer-S~\cite{li2022uniformer} by $2.0$ mIoU, while using less computation complexity. Moreover, compared with UniFormer-B~\cite{li2022uniformer}, our iFormer-S still achieves $0.6$ mIoU improvement with only $1/2$ parameters and nearly $1/3$ FLOPs.

\subsection{Ablation study and visualization}
In this section, we conduct experiments to better understand iFormer. 
All the models are trained for 100 epochs on ImageNet, with the same training setting as described in Sec.~\ref{exp_1k}.

\begin{table}[!b]
    \begin{center}
    \caption{Ablation study of Inception mixer and frequency ramp structure on ImageNet-1K. All the models are trained for 100 epochs.}
    \label{imagenet_ablation}
    \small
    \begin{tabular}{c|ccc|cc|c}
        \toprule 
        \multirow{4}{*}{Mixer} & Attention & MaxPool  & DwConv & \#Param. (M) & FLOPs (G) & Top-1(\%) \\
        \noalign{\smallskip}
        \cline{2-7}
        \noalign{\smallskip}
        ~ & \Checkmark & \XSolidBrush  & \XSolidBrush & 21 & 5.2 & 80.8  \\
        ~ & \Checkmark & \Checkmark & \XSolidBrush   & 20 & 4.9 & 81.0  \\
        ~ & \Checkmark & \Checkmark & \Checkmark   & 20 & 4.8 & 81.2  \\
        \midrule
        \multirow{3}{*}{Structure} & \multicolumn{3}{c|}{\footnotesize{$C_l/C \downarrow , C_h/C \uparrow$ } } & 19 & 4.7 & 80.5 \\
        ~ & \multicolumn{3}{c|}{\footnotesize{$C_l/C  =  C_h/C $ } }                                           & 19 & 4.7 & 80.7 \\
        ~ & \multicolumn{3}{c|}{\footnotesize{$C_l/C \uparrow, C_h/C \downarrow $ }}                           & 20 & 4.8 & 81.2 \\
        \bottomrule
    \end{tabular}
    \end{center}
\end{table}

\begin{figure}[!t]
	\centering
	\subfigure[ 4-$th$ layer ]{
        \label{fig_3branch:a}
        \includegraphics[width=0.46\textwidth]{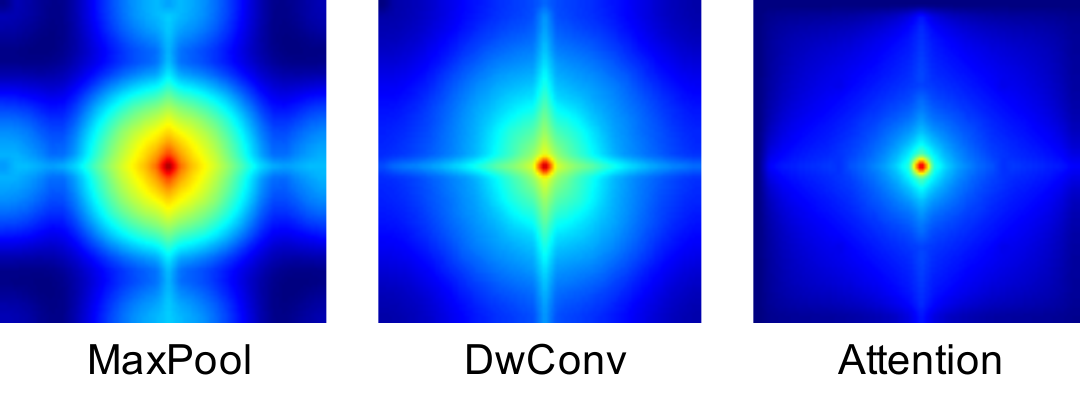}
        }
    \hspace{4mm}
    \subfigure[8-$th$ layer ]{
        \label{fig_3branch:b}
        \includegraphics[width=0.46\textwidth]{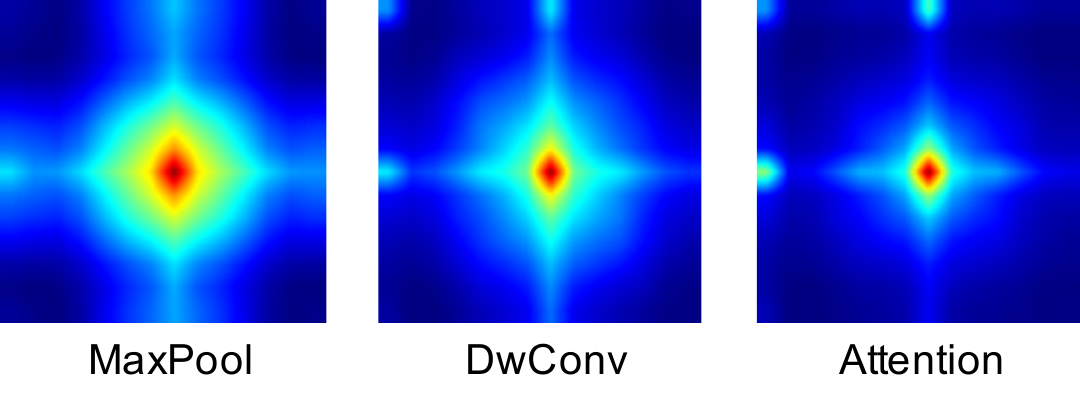}
        }
  	\caption{\textbf{(a) (b) Fourier spectrum of iFormer-S for the MaxPool, DwConv and Attention branches in the Inception mixer.} We can observe that attention mixer tends to reduce high-frequencies, while MaxPool and DwConv enhance them.
  	}
  	\label{fig_3branch}
\end{figure}

\noindent{\textbf{Inception token mixer.}}  
The Inception mixer is proposed to augment the perception capability of ViTs in the frequency spectrum. 
To evaluate the effects of the components in the Inception mixer, we increasingly remove each branch from the full model and then report the results in Table~\ref{imagenet_ablation}, where \Checkmark and \XSolidBrush denote whether or not the corresponding branch is enabled. 
Observably, combining attention with convolution and max-pooling can achieve better accuracy than the attention-only mixer, while using less computation complexity, which implies the effectiveness of Inception Token Mixer. 
To further explore this scheme, Fig.~\ref{fig_3branch} visualizes the Fourier spectrum of the Attention, MaxPool and DwConv branches in Inception mixer. 
We can see the attention mixer has higher concentrations on low frequencies; with the high-frequency mixer, \ie convolution and max-pooling, the model is encouraged to learn high frequency information. 
Overall, these results prove the effectiveness of the Inception mixer for expanding the perception capability of the Transformer in the frequency spectrum.

\noindent{\textbf{Frequency ramp structure.}}  
Previous investigations \cite{raghu2021vision} show  requirement of more local information at lower layers of the Transformer and more  global information at higher layers. 
We accordingly assume that a frequency ramp structure, \ie decreasing dimensions at high-frequency components and increasing dimensions at low-frequency components from lower to higher layers, has a better trade-off between high-frequency and low-frequency components across all layers. In order to justify this hypothesis, we investigate the effects of the channel ratio ($\frac{C_h}{C}$ and $\frac{C_l}{C}$) in Table~\ref{imagenet_ablation}. It can be clearly seen that the model with $C_l/C \uparrow, C_h/C \downarrow $ outperforms the other two models, which is consistent with the previous investigations. 
Hence, this indicates the rationality of the frequency ramp structure and its potential for leaning discriminating vision representations.

\begin{figure}
  \centering
  \includegraphics[width=1.0\textwidth]{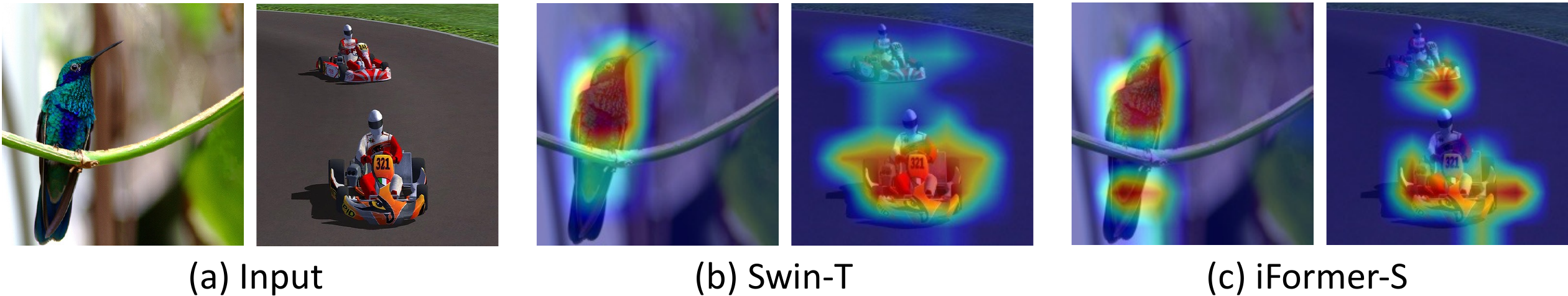}
  \caption{\textbf{\label{grad_cam} Grad-CAM \cite{gradcam} activation maps of Swin-T \cite{liu2021swin} and  iFormer-S trained on ImageNet. }}
\end{figure}

\noindent{\textbf{Visualization.}}  
We visualize the Grad-CAM \cite{gradcam} activation maps of iFormer-S as well as Swin-T \cite{liu2021swin} models trained on ImageNet-1K in Fig. \ref{grad_cam}. It can be seen that compared with Swin,  iFormer can more accurately and completely locate the objects. For example, in the hummingbird image, iFormer skips the branch and accurately attends to the whole bird including the tail.

\section{Conclusion}
\label{concl}

In this paper, we present an Inception Transformer (iFormer), a novel and general Transformer backbone. iFormer adopts a channel splitting mechanism to simply and efficiently couple convolution/max-pooling and self-attention, giving more concentrations on high frequencies and expanding the perception capability of the Transformer in the frequency spectrum. 
Based on the flexible Inception token mixer, we further design a frequency ramp structure, enabling  effective trade-off between high-frequency and low-frequency components across all layers. 
Extensive experiments show that iFormer outperforms representative vision Transformers on image classification, object detection and semantic segmentation, demonstrating the great potential of our iFormer to serve as a general-purpose backbone for computer vision. 
We hope this study will provide valuable insights for the community to design efficient and effective Transformer architectures.  

\paragraph{Limitation.}
One obvious limitation of the proposed iFormer is that it requires manually defined channel ratio in the frequency ramp structure \ie $\frac{C_h}{C}$ and $\frac{C_l}{C}$ for each iFormer block, which needs rich experience to define better on different tasks.
it is not trained on large scale datasets, \eg ImageNet-21K \cite{alexnet}, due to computational constraint, which will be explored in  further. Also, iFormer requires manually defined channel ratio in the frequency ramp structure \ie $\frac{C_h}{C}$ and $\frac{C_l}{C}$ for each iFormer block, which needs rich experience to define better on different tasks.
A straightforward solution would be to use neural architecture search.

\section*{Acknowledgement}
Weihao Yu would like to thank TRC program and GCP research credits for the support of partial computational resources.
{\small
\bibliographystyle{unsrt}
\bibliography{egbib}

\begin{thebibliography}{10}

\bibitem{transformer}
Ashish Vaswani, Noam Shazeer, Niki Parmar, Jakob Uszkoreit, Llion Jones,
  Aidan~N Gomez, {\L}ukasz Kaiser, and Illia Polosukhin.
\newblock Attention is all you need.
\newblock {\em Advances in neural information processing systems}, 30, 2017.

\bibitem{brown2020language}
Tom Brown, Benjamin Mann, Nick Ryder, Melanie Subbiah, Jared~D Kaplan, Prafulla
  Dhariwal, Arvind Neelakantan, Pranav Shyam, Girish Sastry, Amanda Askell,
  et~al.
\newblock Language models are few-shot learners.
\newblock {\em Advances in neural information processing systems},
  33:1877--1901, 2020.

\bibitem{chowdhery2022palm}
Aakanksha Chowdhery, Sharan Narang, Jacob Devlin, Maarten Bosma, Gaurav Mishra,
  Adam Roberts, Paul Barham, Hyung~Won Chung, Charles Sutton, Sebastian
  Gehrmann, et~al.
\newblock Palm: Scaling language modeling with pathways.
\newblock {\em arXiv preprint arXiv:2204.02311}, 2022.

\bibitem{dosovitskiy2020image}
Alexey Dosovitskiy, Lucas Beyer, Alexander Kolesnikov, Dirk Weissenborn,
  Xiaohua Zhai, Thomas Unterthiner, Mostafa Dehghani, Matthias Minderer, Georg
  Heigold, Sylvain Gelly, et~al.
\newblock An image is worth 16x16 words: Transformers for image recognition at
  scale.
\newblock In {\em International Conference on Learning Representations}, 2020.

\bibitem{liu2021swin}
Ze~Liu, Yutong Lin, Yue Cao, Han Hu, Yixuan Wei, Zheng Zhang, Stephen Lin, and
  Baining Guo.
\newblock Swin transformer: Hierarchical vision transformer using shifted
  windows.
\newblock In {\em Proceedings of the IEEE/CVF International Conference on
  Computer Vision}, pages 10012--10022, 2021.

\bibitem{pvt}
Wenhai Wang, Enze Xie, Xiang Li, Deng-Ping Fan, Kaitao Song, Ding Liang, Tong
  Lu, Ping Luo, and Ling Shao.
\newblock Pyramid vision transformer: A versatile backbone for dense prediction
  without convolutions.
\newblock In {\em Proceedings of the IEEE/CVF International Conference on
  Computer Vision}, pages 568--578, 2021.

\bibitem{yu2021metaformer}
Weihao Yu, Mi~Luo, Pan Zhou, Chenyang Si, Yichen Zhou, Xinchao Wang, Jiashi
  Feng, and Shuicheng Yan.
\newblock Metaformer is actually what you need for vision.
\newblock In {\em Proceedings of the IEEE/CVF conference on computer vision and
  pattern recognition}, 2022.

\bibitem{hudson2021generative}
Drew~A Hudson and Larry Zitnick.
\newblock Generative adversarial transformers.
\newblock In {\em International Conference on Machine Learning}, pages
  4487--4499. PMLR, 2021.

\bibitem{arnab2021vivit}
Anurag Arnab, Mostafa Dehghani, Georg Heigold, Chen Sun, Mario Lu{\v{c}}i{\'c},
  and Cordelia Schmid.
\newblock Vivit: A video vision transformer.
\newblock In {\em Proceedings of the IEEE/CVF International Conference on
  Computer Vision}, pages 6836--6846, 2021.

\bibitem{beal2020toward}
Josh Beal, Eric Kim, Eric Tzeng, Dong~Huk Park, Andrew Zhai, and Dmitry
  Kislyuk.
\newblock Toward transformer-based object detection.
\newblock {\em arXiv preprint arXiv:2012.09958}, 2020.

\bibitem{fang2021you}
Yuxin Fang, Bencheng Liao, Xinggang Wang, Jiemin Fang, Jiyang Qi, Rui Wu,
  Jianwei Niu, and Wenyu Liu.
\newblock You only look at one sequence: Rethinking transformer in vision
  through object detection.
\newblock {\em Advances in Neural Information Processing Systems}, 34, 2021.

\bibitem{zheng2021rethinking}
Sixiao Zheng, Jiachen Lu, Hengshuang Zhao, Xiatian Zhu, Zekun Luo, Yabiao Wang,
  Yanwei Fu, Jianfeng Feng, Tao Xiang, Philip~HS Torr, et~al.
\newblock Rethinking semantic segmentation from a sequence-to-sequence
  perspective with transformers.
\newblock In {\em Proceedings of the IEEE/CVF conference on computer vision and
  pattern recognition}, pages 6881--6890, 2021.

\bibitem{xie2021segformer}
Enze Xie, Wenhai Wang, Zhiding Yu, Anima Anandkumar, Jose~M Alvarez, and Ping
  Luo.
\newblock Segformer: Simple and efficient design for semantic segmentation with
  transformers.
\newblock {\em Advances in Neural Information Processing Systems}, 34, 2021.

\bibitem{park2021vision}
Namuk Park and Songkuk Kim.
\newblock How do vision transformers work?
\newblock In {\em International Conference on Learning Representations}, 2021.

\bibitem{bullier2001integrated}
Jean Bullier.
\newblock Integrated model of visual processing.
\newblock {\em Brain research reviews}, 36(2-3):96--107, 2001.

\bibitem{bar2003cortical}
Moshe Bar.
\newblock A cortical mechanism for triggering top-down facilitation in visual
  object recognition.
\newblock {\em Journal of cognitive neuroscience}, 15(4):600--609, 2003.

\bibitem{kauffmann2014neural}
Louise Kauffmann, Stephen Ramano{\"e}l, and Carole Peyrin.
\newblock The neural bases of spatial frequency processing during scene
  perception.
\newblock {\em Frontiers in integrative neuroscience}, 8:37, 2014.

\bibitem{vit}
Alexey Dosovitskiy, Lucas Beyer, Alexander Kolesnikov, Dirk Weissenborn,
  Xiaohua Zhai, Thomas Unterthiner, Mostafa Dehghani, Matthias Minderer, Georg
  Heigold, Sylvain Gelly, et~al.
\newblock An image is worth 16x16 words: Transformers for image recognition at
  scale.
\newblock {\em arXiv preprint arXiv:2010.11929}, 2020.

\bibitem{wang2020high}
Haohan Wang, Xindi Wu, Zeyi Huang, and Eric~P Xing.
\newblock High-frequency component helps explain the generalization of
  convolutional neural networks.
\newblock In {\em Proceedings of the IEEE/CVF Conference on Computer Vision and
  Pattern Recognition}, pages 8684--8694, 2020.

\bibitem{yin2019fourier}
Dong Yin, Raphael Gontijo~Lopes, Jon Shlens, Ekin~Dogus Cubuk, and Justin
  Gilmer.
\newblock A fourier perspective on model robustness in computer vision.
\newblock {\em Advances in Neural Information Processing Systems}, 32, 2019.

\bibitem{xiao2021early}
Tete Xiao, Mannat Singh, Eric Mintun, Trevor Darrell, Piotr Doll{\'a}r, and
  Ross Girshick.
\newblock Early convolutions help transformers see better.
\newblock {\em Advances in Neural Information Processing Systems},
  34:30392--30400, 2021.

\bibitem{li2022uniformer}
Kunchang Li, Yali Wang, Peng Gao, Guanglu Song, Yu~Liu, Hongsheng Li, and
  Yu~Qiao.
\newblock Uniformer: Unified transformer for efficient spatiotemporal
  representation learning.
\newblock {\em arXiv preprint arXiv:2201.04676}, 2022.

\bibitem{xu2021vitae}
Yufei Xu, Qiming Zhang, Jing Zhang, and Dacheng Tao.
\newblock Vitae: Vision transformer advanced by exploring intrinsic inductive
  bias.
\newblock {\em Advances in Neural Information Processing Systems}, 34, 2021.

\bibitem{dai2021coatnet}
Zihang Dai, Hanxiao Liu, Quoc~V Le, and Mingxing Tan.
\newblock Coatnet: Marrying convolution and attention for all data sizes.
\newblock {\em Advances in Neural Information Processing Systems},
  34:3965--3977, 2021.

\bibitem{cvt}
Haiping Wu, Bin Xiao, Noel Codella, Mengchen Liu, Xiyang Dai, Lu~Yuan, and Lei
  Zhang.
\newblock Cvt: Introducing convolutions to vision transformers.
\newblock In {\em Proceedings of the IEEE/CVF International Conference on
  Computer Vision}, pages 22--31, 2021.

\bibitem{xu2021co}
Weijian Xu, Yifan Xu, Tyler Chang, and Zhuowen Tu.
\newblock Co-scale conv-attentional image transformers.
\newblock In {\em Proceedings of the IEEE/CVF International Conference on
  Computer Vision}, pages 9981--9990, 2021.

\bibitem{raghu2021vision}
Maithra Raghu, Thomas Unterthiner, Simon Kornblith, Chiyuan Zhang, and Alexey
  Dosovitskiy.
\newblock Do vision transformers see like convolutional neural networks?
\newblock {\em Advances in Neural Information Processing Systems}, 34, 2021.

\bibitem{deng2009imagenet}
Jia Deng, Wei Dong, Richard Socher, Li-Jia Li, Kai Li, and Li~Fei-Fei.
\newblock Imagenet: A large-scale hierarchical image database.
\newblock In {\em 2009 IEEE conference on computer vision and pattern
  recognition}, pages 248--255. Ieee, 2009.

\bibitem{deit}
Hugo Touvron, Matthieu Cord, Matthijs Douze, Francisco Massa, Alexandre
  Sablayrolles, and Herv{\'e} J{\'e}gou.
\newblock Training data-efficient image transformers \& distillation through
  attention.
\newblock In {\em International Conference on Machine Learning}, pages
  10347--10357. PMLR, 2021.

\bibitem{convnext}
Zhuang Liu, Hanzi Mao, Chao-Yuan Wu, Christoph Feichtenhofer, Trevor Darrell,
  and Saining Xie.
\newblock A convnet for the 2020s.
\newblock {\em arXiv preprint arXiv:2201.03545}, 2022.

\bibitem{lin2014microsoft}
Tsung-Yi Lin, Michael Maire, Serge Belongie, James Hays, Pietro Perona, Deva
  Ramanan, Piotr Doll{\'a}r, and C~Lawrence Zitnick.
\newblock Microsoft coco: Common objects in context.
\newblock In {\em European conference on computer vision}, pages 740--755.
  Springer, 2014.

\bibitem{zhou2017scene}
Bolei Zhou, Hang Zhao, Xavier Puig, Sanja Fidler, Adela Barriuso, and Antonio
  Torralba.
\newblock Scene parsing through ade20k dataset.
\newblock In {\em Proceedings of the IEEE conference on computer vision and
  pattern recognition}, pages 633--641, 2017.

\bibitem{bert}
Jacob Devlin, Ming-Wei Chang, Kenton Lee, and Kristina Toutanova.
\newblock Bert: Pre-training of deep bidirectional transformers for language
  understanding.
\newblock {\em arXiv preprint arXiv:1810.04805}, 2018.

\bibitem{xlnet}
Zhilin Yang, Zihang Dai, Yiming Yang, Jaime Carbonell, Russ~R Salakhutdinov,
  and Quoc~V Le.
\newblock Xlnet: Generalized autoregressive pretraining for language
  understanding.
\newblock {\em Advances in neural information processing systems}, 32, 2019.

\bibitem{roberta}
Yinhan Liu, Myle Ott, Naman Goyal, Jingfei Du, Mandar Joshi, Danqi Chen, Omer
  Levy, Mike Lewis, Luke Zettlemoyer, and Veselin Stoyanov.
\newblock Roberta: A robustly optimized bert pretraining approach.
\newblock {\em arXiv preprint arXiv:1907.11692}, 2019.

\bibitem{gpt}
Alec Radford, Karthik Narasimhan, Tim Salimans, and Ilya Sutskever.
\newblock Improving language understanding by generative pre-training.
\newblock 2018.

\bibitem{gpt3}
Tom Brown, Benjamin Mann, Nick Ryder, Melanie Subbiah, Jared~D Kaplan, Prafulla
  Dhariwal, Arvind Neelakantan, Pranav Shyam, Girish Sastry, Amanda Askell,
  et~al.
\newblock Language models are few-shot learners.
\newblock {\em Advances in neural information processing systems},
  33:1877--1901, 2020.

\bibitem{t2t}
Li~Yuan, Yunpeng Chen, Tao Wang, Weihao Yu, Yujun Shi, Zi-Hang Jiang,
  Francis~EH Tay, Jiashi Feng, and Shuicheng Yan.
\newblock Tokens-to-token vit: Training vision transformers from scratch on
  imagenet.
\newblock In {\em Proceedings of the IEEE/CVF International Conference on
  Computer Vision}, pages 558--567, 2021.

\bibitem{detr}
Nicolas Carion, Francisco Massa, Gabriel Synnaeve, Nicolas Usunier, Alexander
  Kirillov, and Sergey Zagoruyko.
\newblock End-to-end object detection with transformers.
\newblock In {\em European conference on computer vision}, pages 213--229.
  Springer, 2020.

\bibitem{zhu2020deformable}
Xizhou Zhu, Weijie Su, Lewei Lu, Bin Li, Xiaogang Wang, and Jifeng Dai.
\newblock Deformable detr: Deformable transformers for end-to-end object
  detection.
\newblock {\em arXiv preprint arXiv:2010.04159}, 2020.

\bibitem{setr}
Sixiao Zheng, Jiachen Lu, Hengshuang Zhao, Xiatian Zhu, Zekun Luo, Yabiao Wang,
  Yanwei Fu, Jianfeng Feng, Tao Xiang, Philip~HS Torr, et~al.
\newblock Rethinking semantic segmentation from a sequence-to-sequence
  perspective with transformers.
\newblock In {\em Proceedings of the IEEE/CVF conference on computer vision and
  pattern recognition}, pages 6881--6890, 2021.

\bibitem{transunet}
Jieneng Chen, Yongyi Lu, Qihang Yu, Xiangde Luo, Ehsan Adeli, Yan Wang, Le~Lu,
  Alan~L Yuille, and Yuyin Zhou.
\newblock Transunet: Transformers make strong encoders for medical image
  segmentation.
\newblock {\em arXiv preprint arXiv:2102.04306}, 2021.

\bibitem{lecun1998gradient}
Yann LeCun, L{\'e}on Bottou, Yoshua Bengio, and Patrick Haffner.
\newblock Gradient-based learning applied to document recognition.
\newblock {\em Proceedings of the IEEE}, 86(11):2278--2324, 1998.

\bibitem{krizhevsky2012imagenet}
Alex Krizhevsky, Ilya Sutskever, and Geoffrey~E Hinton.
\newblock Imagenet classification with deep convolutional neural networks.
\newblock {\em Advances in neural information processing systems}, 25, 2012.

\bibitem{simonyan2014very}
Karen Simonyan and Andrew Zisserman.
\newblock Very deep convolutional networks for large-scale image recognition.
\newblock {\em arXiv preprint arXiv:1409.1556}, 2014.

\bibitem{szegedy2015going}
Christian Szegedy, Wei Liu, Yangqing Jia, Pierre Sermanet, Scott Reed, Dragomir
  Anguelov, Dumitru Erhan, Vincent Vanhoucke, and Andrew Rabinovich.
\newblock Going deeper with convolutions.
\newblock In {\em Proceedings of the IEEE conference on computer vision and
  pattern recognition}, pages 1--9, 2015.

\bibitem{resnet}
Kaiming He, Xiangyu Zhang, Shaoqing Ren, and Jian Sun.
\newblock Deep residual learning for image recognition.
\newblock In {\em Proceedings of the IEEE conference on computer vision and
  pattern recognition}, pages 770--778, 2016.

\bibitem{alexnet}
Alex Krizhevsky, Ilya Sutskever, and Geoffrey~E Hinton.
\newblock Imagenet classification with deep convolutional neural networks.
\newblock {\em Advances in neural information processing systems}, 25, 2012.

\bibitem{vgg}
Karen Simonyan and Andrew Zisserman.
\newblock Very deep convolutional networks for large-scale image recognition.
\newblock {\em arXiv preprint arXiv:1409.1556}, 2014.

\bibitem{jiang2021all}
Zi-Hang Jiang, Qibin Hou, Li~Yuan, Daquan Zhou, Yujun Shi, Xiaojie Jin, Anran
  Wang, and Jiashi Feng.
\newblock All tokens matter: Token labeling for training better vision
  transformers.
\newblock {\em Advances in Neural Information Processing Systems}, 34, 2021.

\bibitem{tu2022maxvit}
Zhengzhong Tu, Hossein Talebi, Han Zhang, Feng Yang, Peyman Milanfar, Alan
  Bovik, and Yinxiao Li.
\newblock Maxvit: Multi-axis vision transformer.
\newblock {\em arXiv preprint arXiv:2204.01697}, 2022.

\bibitem{chen2021glit}
Boyu Chen, Peixia Li, Chuming Li, Baopu Li, Lei Bai, Chen Lin, Ming Sun, Junjie
  Yan, and Wanli Ouyang.
\newblock Glit: Neural architecture search for global and local image
  transformer.
\newblock In {\em Proceedings of the IEEE/CVF International Conference on
  Computer Vision}, pages 12--21, 2021.

\bibitem{pvtv2}
Wenhai Wang, Enze Xie, Xiang Li, Deng-Ping Fan, Kaitao Song, Ding Liang, Tong
  Lu, Ping Luo, and Ling Shao.
\newblock Pvt v2: Improved baselines with pyramid vision transformer.
\newblock {\em Computational Visual Media}, pages 1--10, 2022.

\bibitem{graham2021levit}
Benjamin Graham, Alaaeldin El-Nouby, Hugo Touvron, Pierre Stock, Armand Joulin,
  Herv{\'e} J{\'e}gou, and Matthijs Douze.
\newblock Levit: a vision transformer in convnet's clothing for faster
  inference.
\newblock In {\em Proceedings of the IEEE/CVF International Conference on
  Computer Vision}, pages 12259--12269, 2021.

\bibitem{zhao2021battle}
Yucheng Zhao, Guangting Wang, Chuanxin Tang, Chong Luo, Wenjun Zeng, and
  Zheng-Jun Zha.
\newblock A battle of network structures: An empirical study of cnn,
  transformer, and mlp.
\newblock {\em arXiv preprint arXiv:2108.13002}, 2021.

\bibitem{ioffe2015batch}
Sergey Ioffe and Christian Szegedy.
\newblock Batch normalization: Accelerating deep network training by reducing
  internal covariate shift.
\newblock In {\em International conference on machine learning}, pages
  448--456. PMLR, 2015.

\bibitem{szegedy2016rethinking}
Christian Szegedy, Vincent Vanhoucke, Sergey Ioffe, Jon Shlens, and Zbigniew
  Wojna.
\newblock Rethinking the inception architecture for computer vision.
\newblock In {\em Proceedings of the IEEE conference on computer vision and
  pattern recognition}, pages 2818--2826, 2016.

\bibitem{szegedy2017inception}
Christian Szegedy, Sergey Ioffe, Vincent Vanhoucke, and Alexander~A Alemi.
\newblock Inception-v4, inception-resnet and the impact of residual connections
  on learning.
\newblock In {\em Thirty-first AAAI conference on artificial intelligence},
  2017.

\bibitem{chollet2017xception}
Fran{\c{c}}ois Chollet.
\newblock Xception: Deep learning with depthwise separable convolutions.
\newblock In {\em Proceedings of the IEEE conference on computer vision and
  pattern recognition}, pages 1251--1258, 2017.

\bibitem{mamalet2012simplifying}
Franck Mamalet and Christophe Garcia.
\newblock Simplifying convnets for fast learning.
\newblock In {\em International Conference on Artificial Neural Networks},
  pages 58--65. Springer, 2012.

\bibitem{sandler2018mobilenetv2}
Mark Sandler, Andrew Howard, Menglong Zhu, Andrey Zhmoginov, and Liang-Chieh
  Chen.
\newblock Mobilenetv2: Inverted residuals and linear bottlenecks.
\newblock In {\em Proceedings of the IEEE conference on computer vision and
  pattern recognition}, pages 4510--4520, 2018.

\bibitem{resnetrsb}
Ross Wightman, Hugo Touvron, and Herv{\'e} J{\'e}gou.
\newblock Resnet strikes back: An improved training procedure in timm.
\newblock {\em arXiv preprint arXiv:2110.00476}, 2021.

\bibitem{yang2021focal}
Jianwei Yang, Chunyuan Li, Pengchuan Zhang, Xiyang Dai, Bin Xiao, Lu~Yuan, and
  Jianfeng Gao.
\newblock Focal self-attention for local-global interactions in vision
  transformers.
\newblock {\em arXiv preprint arXiv:2107.00641}, 2021.

\bibitem{dong2021cswin}
Xiaoyi Dong, Jianmin Bao, Dongdong Chen, Weiming Zhang, Nenghai Yu, Lu~Yuan,
  Dong Chen, and Baining Guo.
\newblock Cswin transformer: A general vision transformer backbone with
  cross-shaped windows.
\newblock {\em arXiv preprint arXiv:2107.00652}, 2021.

\bibitem{lu2021container}
Jiasen Lu, Roozbeh Mottaghi, Aniruddha Kembhavi, et~al.
\newblock Container: Context aggregation networks.
\newblock {\em Advances in Neural Information Processing Systems}, 34, 2021.

\bibitem{zhang2022vitaev2}
Qiming Zhang, Yufei Xu, Jing Zhang, and Dacheng Tao.
\newblock Vitaev2: Vision transformer advanced by exploring inductive bias for
  image recognition and beyond.
\newblock {\em arXiv preprint arXiv:2202.10108}, 2022.

\bibitem{regnet}
Ilija Radosavovic, Raj~Prateek Kosaraju, Ross Girshick, Kaiming He, and Piotr
  Doll{\'a}r.
\newblock Designing network design spaces.
\newblock In {\em Proceedings of the IEEE/CVF Conference on Computer Vision and
  Pattern Recognition}, pages 10428--10436, 2020.

\bibitem{srinivas2021bottleneck}
Aravind Srinivas, Tsung-Yi Lin, Niki Parmar, Jonathon Shlens, Pieter Abbeel,
  and Ashish Vaswani.
\newblock Bottleneck transformers for visual recognition.
\newblock In {\em Proceedings of the IEEE/CVF conference on computer vision and
  pattern recognition}, pages 16519--16529, 2021.

\bibitem{loshchilov2016sgdr}
Ilya Loshchilov and Frank Hutter.
\newblock Sgdr: Stochastic gradient descent with warm restarts.
\newblock {\em arXiv preprint arXiv:1608.03983}, 2016.

\bibitem{touvron2021going}
Hugo Touvron, Matthieu Cord, Alexandre Sablayrolles, Gabriel Synnaeve, and
  Herv{\'e} J{\'e}gou.
\newblock Going deeper with image transformers.
\newblock In {\em Proceedings of the IEEE/CVF International Conference on
  Computer Vision}, pages 32--42, 2021.

\bibitem{timm}
Ross Wightman.
\newblock Pytorch image models.
\newblock \url{https://github.com/rwightman/pytorch-image-models}, 2019.

\bibitem{tan2019efficientnet}
Mingxing Tan and Quoc Le.
\newblock Efficientnet: Rethinking model scaling for convolutional neural
  networks.
\newblock In {\em International conference on machine learning}, pages
  6105--6114. PMLR, 2019.

\bibitem{tan2021efficientnetv2}
Mingxing Tan and Quoc Le.
\newblock Efficientnetv2: Smaller models and faster training.
\newblock In {\em International Conference on Machine Learning}, pages
  10096--10106. PMLR, 2021.

\bibitem{he2017mask}
Kaiming He, Georgia Gkioxari, Piotr Doll{\'a}r, and Ross Girshick.
\newblock Mask r-cnn.
\newblock In {\em Proceedings of the IEEE international conference on computer
  vision}, pages 2961--2969, 2017.

\bibitem{chu2021twins}
Xiangxiang Chu, Zhi Tian, Yuqing Wang, Bo~Zhang, Haibing Ren, Xiaolin Wei,
  Huaxia Xia, and Chunhua Shen.
\newblock Twins: Revisiting the design of spatial attention in vision
  transformers.
\newblock {\em Advances in Neural Information Processing Systems}, 34, 2021.

\bibitem{zhang2021multi}
Pengchuan Zhang, Xiyang Dai, Jianwei Yang, Bin Xiao, Lu~Yuan, Lei Zhang, and
  Jianfeng Gao.
\newblock Multi-scale vision longformer: A new vision transformer for
  high-resolution image encoding.
\newblock In {\em Proceedings of the IEEE/CVF International Conference on
  Computer Vision}, pages 2998--3008, 2021.

\bibitem{mmdetection}
Kai Chen, Jiaqi Wang, Jiangmiao Pang, Yuhang Cao, Yu~Xiong, Xiaoxiao Li,
  Shuyang Sun, Wansen Feng, Ziwei Liu, Jiarui Xu, Zheng Zhang, Dazhi Cheng,
  Chenchen Zhu, Tianheng Cheng, Qijie Zhao, Buyu Li, Xin Lu, Rui Zhu, Yue Wu,
  Jifeng Dai, Jingdong Wang, Jianping Shi, Wanli Ouyang, Chen~Change Loy, and
  Dahua Lin.
\newblock {MMDetection}: Open mmlab detection toolbox and benchmark.
\newblock {\em arXiv preprint arXiv:1906.07155}, 2019.

\bibitem{kirillov2019panoptic}
Alexander Kirillov, Ross Girshick, Kaiming He, and Piotr Doll{\'a}r.
\newblock Panoptic feature pyramid networks.
\newblock In {\em Proceedings of the IEEE/CVF Conference on Computer Vision and
  Pattern Recognition}, pages 6399--6408, 2019.

\bibitem{mmseg2020}
MMSegmentation Contributors.
\newblock {MMSegmentation}: Openmmlab semantic segmentation toolbox and
  benchmark.
\newblock \url{https://github.com/open-mmlab/mmsegmentation}, 2020.

\bibitem{gradcam}
Ramprasaath~R Selvaraju, Michael Cogswell, Abhishek Das, Ramakrishna Vedantam,
  Devi Parikh, and Dhruv Batra.
\newblock Grad-cam: Visual explanations from deep networks via gradient-based
  localization.
\newblock In {\em Proceedings of the IEEE international conference on computer
  vision}, pages 618--626, 2017.

\bibitem{xiao2018unified}
Tete Xiao, Yingcheng Liu, Bolei Zhou, Yuning Jiang, and Jian Sun.
\newblock Unified perceptual parsing for scene understanding.
\newblock In {\em Proceedings of the European Conference on Computer Vision
  (ECCV)}, pages 418--434, 2018.

\bibitem{huang2021shuffle}
Zilong Huang, Youcheng Ben, Guozhong Luo, Pei Cheng, Gang Yu, and Bin Fu.
\newblock Shuffle transformer: Rethinking spatial shuffle for vision
  transformer.
\newblock {\em arXiv preprint arXiv:2106.03650}, 2021.

\end{thebibliography}
}

\clearpage

\appendix

\section{Appendix}
\textbf{Potential Impacts.} The study introduces a general vision Transformer, \ie iFormer, which can be used  on different vision tasks, \eg image classification, object detection and semantic segmentation. iFormer has no direct negative societal impact. However, we will realize that iFormer as a general-purpose backbone can be used for harmful applications such as illegal face recognition.

\subsection{Results on semantic segmentation}

\paragraph{Setup.} We further evaluate the generality of iFormer on semantic segmentation with the Upernet~\cite{xiao2018unified} framework. Following the training settings in Swin~\cite{liu2021swin},  the model is trained for 160K iterations with a batch size of 16. For training, we use AdamW optimizer with an initial learning rate $6 \times 10^{-5}$. All experiments are implemented on mmsegmentation~\cite{mmseg2020}  codebase.

\paragraph{Results.} Table~\ref{coco_sota_seg_uper} shows the mIoU and MS mIoU results of different backbones based on the UperNet~\cite{xiao2018unified} framework. From these results, it can be seen that our iFormer achieves 48.4 mIoU and 48.8 MS mIoU, consistently surpassing previous backbones on this task. For instance, our iFormer-S outperforms the Swin-T~\cite{liu2021swin} by 3.9 mIoU while using fewer parameters. Compered with UniFormer-S~\cite{li2022uniformer}, iFormer-S still achieves 0.8 mIoU improvement.

\begin{table}[!h]
    \begin{center}
    \caption{Semantic segmentation with semantic UperNet \cite{xiao2018unified} on ADE20K \cite{zhou2017scene}. The FLOPs are measured at resolution 512$\times$2048.}
     \label{coco_sota_seg_uper}
    \begin{tabular}{l|c|c|c|c}
        \toprule 
         \multirow{2}{*}{Method} & \#Param. & FLOPs & mIoU & MS mIoU \\
        ~ & (M)  & (G) & (\%) & (\%)  \\
        \midrule
        TwinsP-S \cite{chu2021twins}       & 55 & 919 & 46.2 & 47.5 \\
        Twins-S \cite{chu2021twins}        & 54 & 901 & 46.2 & 47.1 \\
        Swin-T \cite{liu2021swin}          & 60 & 945 & 44.5 & 45.8 \\
        Focal-T \cite{yang2021focal}       & 62 & 998 & 45.8 & 47.0 \\
        Shuffle-T \cite{huang2021shuffle}                        & 60 & 949 & 46.6 & 47.8 \\
        UniFormer-S$_{h32}$ \cite{li2022uniformer}               & 52 & 955 & 47.0 & 48.5 \\
        UniFormer-S \cite{li2022uniformer}                       & 52 & 1008 & 47.6 & 48.5 \\
        \rowcolor{gray!15} \textbf{iFormer-S}          & \textbf{49} & \textbf{938} & \textbf{48.4} & \textbf{48.8} \\
        \bottomrule
    \end{tabular}
    \end{center}
\end{table}

\subsection{Visualization}

\subsubsection{Fourier spectrum of different layers}

\begin{figure}[!b]
	\centering
	\subfigure[6-$th$ layer ]{
        \label{fig_fft_layer:a}
        \includegraphics[width=0.3\textwidth]{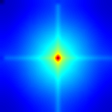}
        }
    \subfigure[12-$th$ layer ]{
        \label{fig_fft_layer:b}
        \includegraphics[width=0.3\textwidth]{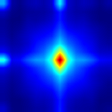}
        }
    \subfigure[18-$th$ layer ]{
        \label{fig_fft_layer:c}
        \includegraphics[width=0.3\textwidth]{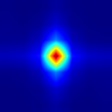}
        }
    \vspace{-2mm}
  	\caption{\textbf{(a) Fourier spectrum of 6-$th$, 12-$th$ and 18-$th$ layers for the iFormer-S.} 
  	}
  	\label{fig_fft_layer}
\end{figure}

\begin{figure}[!t]
  \centering
  \includegraphics[width=0.83\textwidth]{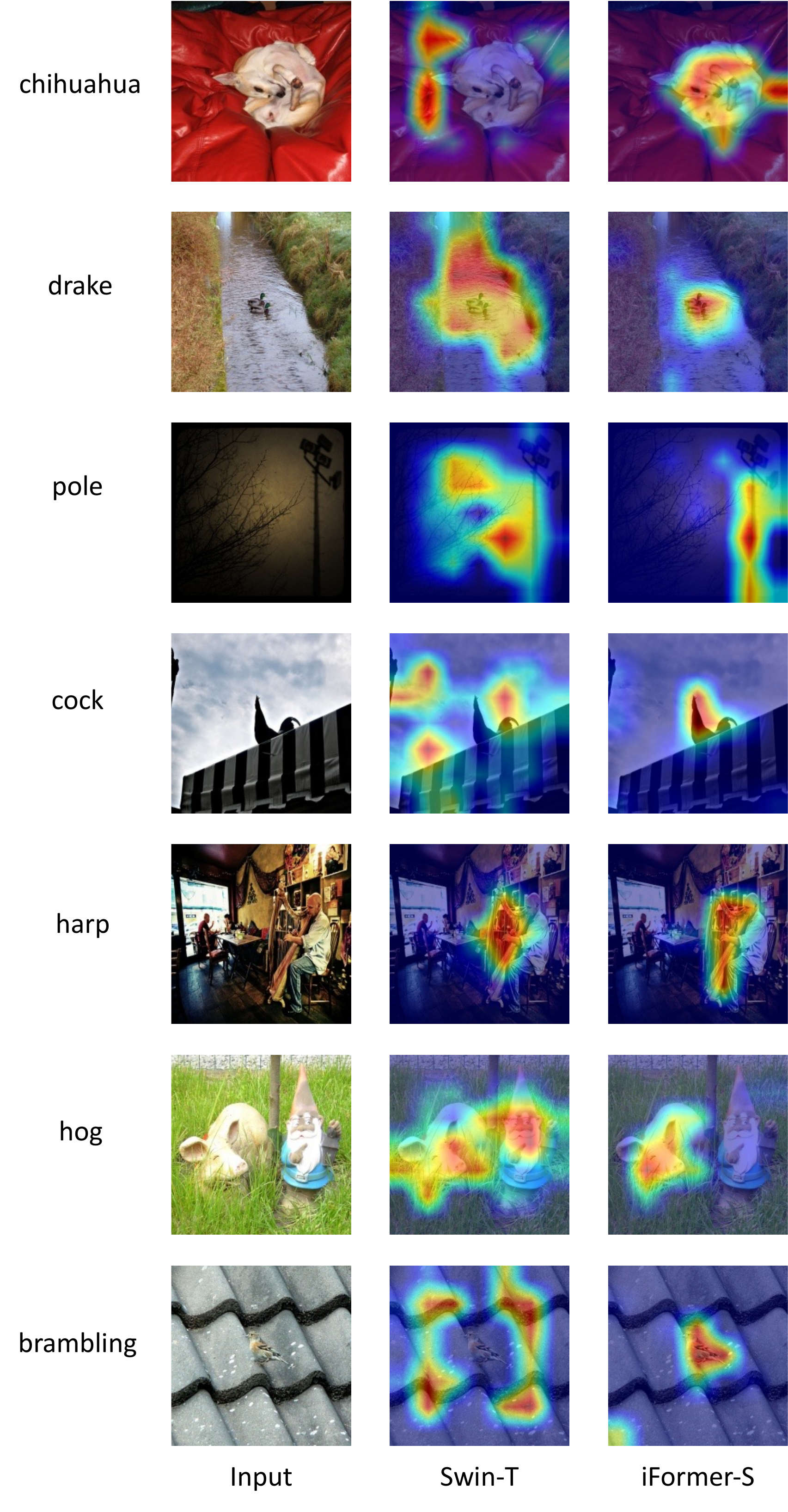}
  \vspace{-3mm}
  \caption{\textbf{\label{grad_cam_app} Grad-CAM \cite{gradcam} activation maps of Swin-T \cite{liu2021swin} and  iFormer-S trained on ImageNet. }}
  \vspace{-15pt}
\end{figure}

Frequency ramp structure plays an important role in iFormer, which is designed to learn hierarchical representations, \ie more high-frequency signals at lower layers of the Transformer and more low-frequency signals at higher layers. In order to justify this hypothesis, we visualize the Fourier spectrum of feature maps for different iFormer layers in Fig~\ref{fig_fft_layer}. We can see that iFormer captures more high-frequency components at 6-$th$ layer and more low-frequency information at 18-$th$ layer. Moreover, the high-frequency information gradually decreases from 6-$th$ layer to 18-$th$ layer, and low-frequency information does the opposite. Hence, these results indicate our iFormer can effectively trade-off high- and low-frequency components across all layers.

\subsubsection{CAM}

We futher show more examples of the Grad-CAM \cite{gradcam} activation maps of iFormer-S as well as Swin-T \cite{liu2021swin} models trained on ImageNet-1K in Fig. \ref{grad_cam_app}. These examples indicate that compared with Swin,  iFormer can more accurately and completely attend to the key objects. Taking the hog picture as example, iFormer locates the hog accurately but Swin also locates irrelevant part.

\subsection{Configurations of iFormers}

In this work, three variants of iFormer are used for a fair comparison under computation configurations, \ie iFormer-S, iFormer-B and iFormer-L. Table~\ref{table_configurations} shows their detailed configurations. Following Swin \cite{liu2021swin}, iFormer adopts 4-stage architecture with $\frac{H}{4} \times \frac{W}{4}$, $\frac{H}{8} \times \frac{W}{8}$, $\frac{H}{16} \times \frac{W}{16}$, $\frac{H}{32} \times \frac{W}{32}$ input sizes, where $H$ and $W$ are the width and height of the input image. In each iFormer block, $C_h/C$ and $C_l/C$ are used to balance the high-frequency and low frequency components. As shown in Table~\ref{table_configurations}, $C_h/C$ gradually decreases from shallow to deep layers, while $C_l/C$ gradually increases. iFormer block uses depthwise convolution and max-pooling as high-frequency mixers. We set the kernel sizes of depthwise convolution and max-pooling to $3 \times 3$.

\begin{table}[!h]
    \begin{center}
    \caption{ Configurations of the variants of iFormer. Pool stride denotes the stride of the pooling and upsample layers in attention branch. The FLOPs are measured at resolution $224 \times 224$.}
    \label{table_configurations}
    \resizebox{1.0\linewidth}{!}{
    \footnotesize{
        \begin{tabular}{c|c|c|c|c}
        \toprule 
        Stage & Layer & iFormer-S & iFormer-B & iFormer-L \\
        \midrule
        \multirow{5}{*}{1} & \makecell[c]{Patch \\ Embedding} & \makecell[c]{$ 3 \times 3, \mathrm{stride}~2, 48 $ \\ $ 3 \times 3, \mathrm{stride}~2, 96 $} & \makecell[c]{ $3 \times 3, \mathrm{stride}~2, 48 $ \\ $ 3 \times 3, \mathrm{stride}~2, 96 $ }& \makecell[c]{ $  3 \times 3, \mathrm{stride}~2, 48 $ \\ $ 3 \times 3, \mathrm{stride}~2, 96 $ } \\
        \cmidrule{2-5}
        ~ & \makecell[c]{iFormer \\ Block} & $\begin{bmatrix} C_h/h = 2/3 \\ C_l/h = 1/3 \\ \mathrm{pool~stride}~2\end{bmatrix} \times 3$ & $\begin{bmatrix} C_h/h = 2/3 \\ C_l/h = 1/3 \\ \mathrm{pool~stride}~2\end{bmatrix} \times 4 $ & $\begin{bmatrix} C_h/h = 2/3 \\ C_l/h = 1/3 \\ \mathrm{pool~stride}~2\end{bmatrix} \times 4 $  \\
        \midrule
        \multirow{5}{*}{2} & \makecell[c]{Patch \\ Embedding} & $ 2 \times 2,  \mathrm{stride}~2, 192$ &  $ 2 \times 2,  \mathrm{stride}~2, 192$ & $ 2 \times 2,  \mathrm{stride}~2, 192$  \\
        \cmidrule{2-5}
        ~ & \makecell[c]{iFormer \\ Block} & $ \begin{bmatrix} C_h/h = 1/2 \\ C_l/h = 1/2 \\ \mathrm{pool~stride}~2\end{bmatrix} \times 3$  & $ \begin{bmatrix} C_h/h = 1/2 \\ C_l/h = 1/2 \\ \mathrm{pool~stride}~2\end{bmatrix} \times 6$ & $ \begin{bmatrix} C_h/h = 1/2 \\ C_l/h = 1/2 \\ \mathrm{pool~stride}~2\end{bmatrix} \times 6$  \\
        \midrule
        \multirow{5}{*}{3} & \makecell[c]{Patch \\ Embedding} & $ 2 \times 2,  \mathrm{stride}~2, 320$ & $ 2 \times 2,  \mathrm{stride}~2, 384$ & $ 2 \times 2,  \mathrm{stride}~2, 448$  \\
        \cmidrule{2-5}
        ~ & \makecell[c]{iFormer \\ Block} & $ \begin{bmatrix} C_h/h = 3/10 \to 1/10 \\ C_l/h = 7/10 \to 9/10 \\ \mathrm{pool~stride}~1\end{bmatrix} \times 9$ & $ \begin{bmatrix} C_h/h = 4/12 \to 2/12 \\ C_l/h = 8/12 \to 10/12 \\ \mathrm{pool~stride}~1\end{bmatrix} \times 14$ & $ \begin{bmatrix} C_h/h = 4/14 \to 2/14 \\ C_l/h = 10/14 \to 12/14 \\ \mathrm{pool~stride}~1\end{bmatrix} \times 18$  \\
        \midrule
        \multirow{5}{*}{4} & \makecell[c]{Patch \\ Embedding} & $ 2 \times 2,  \mathrm{stride}~2, 384$ & $ 2 \times 2,  \mathrm{stride}~2, 512$ & $ 2 \times 2,  \mathrm{stride}~2, 640$  \\
        \cmidrule{2-5}
        ~ & \makecell[c]{iFormer \\ Block} & $ \begin{bmatrix} C_h/h = 1/12 \\ C_l/h = 11/12 \\ \mathrm{pool~stride}~1\end{bmatrix} \times 3$  & $ \begin{bmatrix} C_h/h = 1/16 \\ C_l/h = 15/16 \\ \mathrm{pool~stride}~1\end{bmatrix} \times 6$ & $ \begin{bmatrix} C_h/h = 1/20 \\ C_l/h = 19/20 \\ \mathrm{pool~stride}~1\end{bmatrix} \times 8$  \\
        \midrule
        \multicolumn{2}{c|}{\#Param. (M) }  & 20 & 48 & 87 \\
        \midrule
        \multicolumn{2}{c|}{FLOPs (G) } & 4.8 & 9.4 & 14.0 \\
        \bottomrule
    \end{tabular}
    }}
    \end{center}
\end{table}

\end{document}